\PassOptionsToPackage{table}{xcolor}
\RequirePackage{fix-cm}
\documentclass[10pt,letterpaper]{evolife_report}
\newif\ifreportlayout\reportlayouttrue
\usepackage{amsmath,amssymb,amsfonts}
\usepackage{graphicx,booktabs,multirow,tabularx,array,adjustbox}
\usepackage[table]{xcolor}
\usepackage{algorithm}
\usepackage{algpseudocode}
\usepackage{pifont}
\usepackage{url}
\usepackage{bm}
\usepackage{hyperref}
\usepackage[most]{tcolorbox}
\definecolor{EvoNavy}{HTML}{152C43}
\definecolor{EvoTeal}{HTML}{008B7C}
\definecolor{EvoBlue}{HTML}{0077BB}
\definecolor{EvoOrange}{HTML}{D56729}
\definecolor{EvoTealTint}{HTML}{E8F5F2}
\definecolor{EvoBlueTint}{HTML}{EAF4FB}
\definecolor{EvoOrangeTint}{HTML}{FFF1E7}
\definecolor{EvoGreyTint}{HTML}{F2F6F9}
\definecolor{EvoRule}{HTML}{CBD9E2}
\DeclareRobustCommand{\method}{\textnormal{\fontfamily{lmr}\selectfont\scshape EvoLife}}
\newcommand{\tool}[1]{\textsf{#1}}
\newcommand{\figref}[1]{Fig.~\ref{#1}}
\newcommand{\tabref}[1]{Table~\ref{#1}}

\newtcolorbox{mechanismbox}[1]{enhanced,breakable,colback=EvoTealTint,colframe=EvoTealTint,boxrule=0pt,arc=2mm,borderline west={2pt}{0pt}{EvoTeal},title={#1},fonttitle=\sffamily\bfseries,coltitle=EvoNavy,colbacktitle=EvoTealTint,left=7pt,right=7pt,top=4pt,bottom=5pt}
\newtcolorbox{evidencebox}[1]{enhanced,breakable,colback=EvoGreyTint,colframe=EvoRule,boxrule=.4pt,arc=1.4mm,title={#1},fonttitle=\sffamily\bfseries,coltitle=EvoNavy,colbacktitle=EvoGreyTint}
\hypersetup{colorlinks=true,linkcolor=EvoTeal,citecolor=EvoTeal,urlcolor=EvoBlue,pdftitle={Ask to Evolve: A Unified Harness for Lifelong Embodied Navigation with Hybrid Actions and Persistent Memory},pdfauthor={Kai Sheng et al.}}
\setcitestyle{numbers,square,comma}
\title{Ask to Evolve: A Unified Harness for Lifelong Embodied Navigation with Hybrid Action Space and Persistent Memory}
\author{Kai Sheng}
\author{Liuyi Wang$^{\dagger}$}
\author{Jinlong Li}
\author{Haojie Dai}
\author{Chengju Liu}
\author{Qijun Chen$^{\dagger}$}
\affiliation{School of Electronic and Information Engineering, Tongji University, Shanghai, China}
\astradata[Project Page]{\href{https://kai-sheng-caesar.github.io/EvoLife-Web/}{Project Website}}
\projectlogos{%
  \parbox[b]{0.7\textwidth}{%
    \raggedleft
    \fontsize{8}{10}\selectfont
    \itshape
    ``We can only see a short distance ahead,
    but we can see plenty there that needs to be done.''\par
    \vspace{2pt}
    {\normalfont
    --- Alan M. Turing, 1950~\citep{turing1950}}%
  }%
}
\date{October 2026}
\abstract{Giving embodied agents the freedom to act autonomously, initiate interactions, and understand their surroundings through experience is a step toward bringing them to life in the physical world. Realizing this vision requires action, interaction, and memory to work together, so that each encounter can inform current decisions and contribute experience for future tasks. We present \method{}, a unified navigation harness that enables a multimodal embodied agent to choose how to move, when to ask, and what experience to carry forward. Its hybrid action space supports waypoint navigation, local discrete adjustments, pixel-based navigation, and object-directed navigation. Active interaction allows the agent to seek clarification or assistance when it encounters ambiguity or difficulty, while persistent spatial memory binds observations and dialogue to places, retaining useful experience to guide subsequent visits. We evaluate eight benchmarks across four task families: vision-language navigation, object navigation, interactive instance navigation, and continuous multi-task navigation. 
With an average of only 1.87 active interactions per episode, our method achieves a first-visit success rate (SR) of 100\% across four 100-episode sets spanning vision-language navigation and object navigation.
Complete-sequence SR reaches 98\% on 50 LHPR-VLN sequences and 94\% on 50 GOAT-Bench sequences. In the R2R revisit setting, success weighted by path length (SPL) increases from 82.23\% to 93.31\% with retained experience. Multi-day physical navigation further demonstrates exploration, proactive help-seeking, and revisiting remembered places. Together, these results advance the vision of bringing embodied agents to life through autonomous action, proactive interaction, and the accumulation and reuse of experience.

}
\begin{document}
\maketitle
\begingroup
\renewcommand{\thefootnote}{\fnsymbol{footnote}}
\footnotetext[2]{Corresponding authors.}
\endgroup
\begin{figure}[H]
\centering
\includegraphics[width=.99\linewidth]{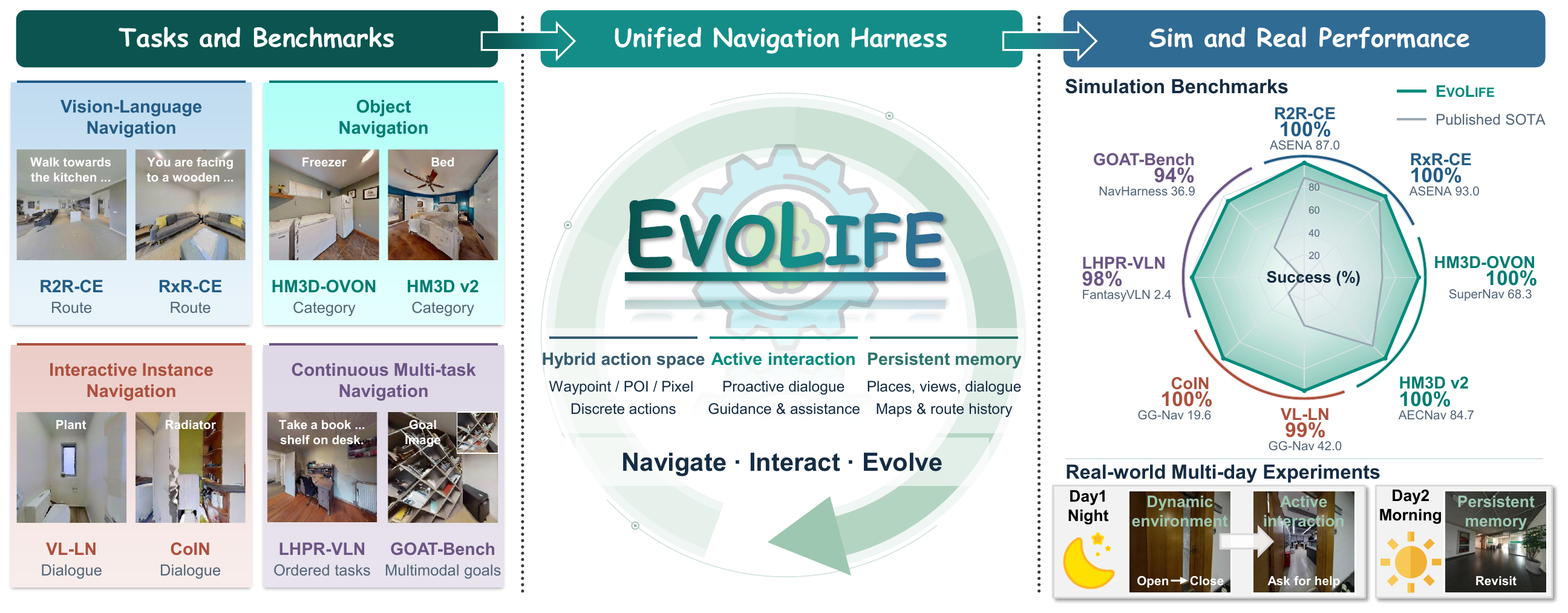}
\caption{\method{} takes a step toward bringing embodied agents to life through autonomous action, proactive interaction, and the accumulation of experience. The left panel presents four task families: vision-language navigation~\cite{r2r2018,rxr2020,vlnce2020}, object navigation~\cite{hm3d2021,hm3dsem2023,ovon2024}, interactive instance navigation~\cite{vlln2025,coin2025}, and continuous multi-task navigation~\cite{lhpr2025,goat2024}. The center shows the unified harness, connecting a hybrid action space, active interaction, and persistent spatial memory. The right panel reports success rate (SR) across eight simulation benchmarks and illustrates multi-day physical navigation, demonstrating the agent's ability to respond to environmental changes, proactively seek assistance, and revisit remembered places.}
\label{fig:teaser}
\end{figure}

\clearpage
\bibliographystyle{references/evolife_ieee}
\section{Introduction}
\label{sec:introduction}

Lifelong embodied navigation calls for agents that can take initiative in how they act, interact, and accumulate experience. An agent should be able to choose how to pursue a goal according to its environment and task requirements, ask questions when it encounters ambiguity, seek assistance when progress becomes difficult, and carry what it learns into future tasks. These capabilities give concrete meaning to the vision of bringing embodied agents to life in the physical world. Through sustained engagement with their surroundings, agents can connect what they perceive with what they do, use interaction to develop their understanding, and build experience that remains useful beyond a single encounter.

People navigating unfamiliar environments naturally ask for clarification, request directions, and seek help when needed. Embodied agents need the freedom to take the same initiative. A question can clarify an instruction, identify a useful continuation, or obtain assistance with a physical obstacle. The value of that interaction can also extend beyond the immediate task: an answer about a place, together with the observations and actions that follow, can become experience for a later visit. This motivates the central idea of {\bfseries\itshape Ask to Evolve}: an agent actively asks to improve its understanding, acts on the resulting information, and retains useful experience to guide future behavior.

Embodied navigation provides concrete settings for studying this process. Vision-language navigation connects linguistic instructions with visual observations and navigation decisions, including movement through continuous environments~\cite{r2r2018,rxr2020,vlnce2020,dai2026dped}. Interactive navigation introduces dialogue as a source of clarification and assistance~\cite{cvdn2020,hanna2019,orion2024}, while iterative and lifelong navigation make earlier encounters relevant to subsequent tasks~\cite{ivln2023,goat2024}. Systematic real-world evaluations have documented a substantial gap between simulated and physical navigation performance~\cite{wang2026comprehensive,wang2025rethinking}. This gap highlights the importance of coordinating perception, decision-making, and execution, together with the ability to obtain assistance and reuse experience in real environments.

Existing methods offer useful foundations. Topological planners connect visual waypoints with continuous control~\cite{etpnav2025}, and language-model navigators integrate observations, reasoning, and actions through execution interfaces~\cite{navgpt2024,navgpt32026}. Memory-based systems retain observations and behavior, while navigation harnesses preserve maps and task records across sessions~\cite{memoir2026,navharness2026,harnessvln2026}. These advances motivate a unified systems question: {\bfseries\itshape how can an embodied agent keep multiple ways of moving, actively requested assistance, and remembered experience available within the same decision loop?} The connection matters because an interaction may change the next movement, revise the interpretation of a place, and contribute knowledge that is useful during a subsequent task.

We introduce \method{}, a unified harness for lifelong embodied navigation with a hybrid action space, active interaction, and persistent spatial memory. Its action space supports four complementary modes: waypoint navigation toward a selected spatial location, local discrete adjustments for fine control, pixel-based navigation toward a selected image location, and object-directed navigation toward an identified target. These modes give the agent flexibility in expressing its movement intention. It can select a spatial waypoint, make a short adjustment, indicate a visible location, or approach an object, according to the task and available observations. Execution returns the actual outcome, allowing the agent to inspect its progress and choose its next action. \figref{fig:overview} presents the overall framework.

Active interaction is part of this ongoing decision process. The agent chooses when to ask and formulates its question using the current task, observations, and relevant memory. It can request clarification when an instruction is ambiguous, ask for guidance when the continuation is uncertain, or seek practical assistance when the environment prevents progress. After receiving a response, the agent decides how to act and uses subsequent observations and execution feedback to assess the outcome. Interaction therefore gives the agent an active means of obtaining information and assistance while remaining connected to movement and the experience generated through execution.

Persistent spatial memory allows these experiences to accumulate across tasks. Place records bind locations to descriptions, observations, historical images, and dialogue. Information acquired through an interaction can update a place description, support a later navigation decision, or help interpret a familiar environment under changed conditions. New observations can, in turn, prompt revisions to existing records. 
This continuous accumulation and refinement of spatial knowledge allows the agent's experience to evolve across successive tasks. By creating, retrieving, updating, and reusing historical memories, the agent can not only acquire new knowledge from current encounters but also leverage evolving experience to guide future decisions.

We evaluate \method{} on eight navigation benchmarks across four task families: vision-language navigation~\cite{r2r2018,rxr2020,vlnce2020}, object navigation~\cite{hm3d2021,hm3dsem2023,ovon2024}, interactive instance navigation~\cite{vlln2025,coin2025}, and continuous multi-task navigation~\cite{lhpr2025,goat2024}. The main study includes six sets of 100 single-task episodes and two sets of 50 continuous sequences.
With an average of only 1.87 active interactions per episode, our method achieves a success rate (SR) of 100\% across four 100-episode sets spanning vision-language navigation and object navigation.
Complete-sequence SR reaches 98\% on LHPR-VLN and 94\% on GOAT-Bench. In the R2R revisit setting, success weighted by path length (SPL) increases from 82.23\% to 93.31\% with retained experience. Ablation analyses further examine interaction and experience reuse, while multi-day physical experiments demonstrate exploration, proactive help-seeking, and revisiting remembered places.


Our contributions are threefold:
\begin{enumerate}
    \item \textbf{Hybrid action space.}
    We propose \method{}, which enables an embodied agent to select among four navigation modes and adapt subsequent decisions based on execution feedback.

    \item \textbf{Active interaction.}
    We integrate proactive interaction into the navigation loop, allowing the agent to autonomously decide when to ask, generate clarification or assistance requests, and incorporate responses into navigation decisions.

    \item \textbf{Persistent memory.}
    We develop a dynamic long-term memory mechanism that associates observations, descriptions, and dialogue with spatial locations, enabling explicit memory updates and continuous experience accumulation across tasks.
\end{enumerate}

Together, extensive benchmark evaluations and multi-day physical experiments advance the vision of bringing embodied agents to life through autonomous action, proactive interaction, and accumulated experience that guides future encounters.

\section{Related Work}
\label{sec:related}

\subsection{Embodied Navigation}
Embodied navigation connects visual perception and language understanding with executable movement~\cite{wang2024causal,wang2026magic}. R2R established instruction-guided indoor navigation, RxR expanded instruction diversity through detailed multilingual descriptions, and continuous VLN extended execution beyond discrete viewpoint graphs~\cite{r2r2018,rxr2020,vlnce2020}. Surveys and systematic real-world evaluations organize VLN through action and model paradigms and identify challenges in transferring navigation capabilities from simulation to physical environments~\cite{wang2026comprehensive}. These developments make the connection between decision-making and execution a central issue in embodied navigation.

Existing approaches organize this connection in different ways. ETPNav combines online waypoint topology, cross-modal planning, and obstacle-aware execution~\cite{etpnav2025}. NavGPT and NavGPT-3 explore explicit reasoning and interfaces connecting reasoning, action, and runtime execution~\cite{navgpt2024,navgpt32026}. NaVid, StreamVLN, and NaVILA connect visual history and language with executable behavior~\cite{navid2024,streamvln2025,navila2025}. P2DNav uses panoramic direction selection and downview grounding, NavJev introduces action-specific visual compression, and CLASH coordinates reactive and reflective planning~\cite{p2dnav2026,navjev2026,clash2025}. Correctable landmark discovery and disentangled navigation reasoning provide further mechanisms for interpreting observations and selecting actions~\cite{console2024,navcot2025,yu2026correctnav}.

These studies highlight the importance of how movement intentions are expressed and executed. \method{} integrates waypoint navigation, local discrete adjustments, pixel-based navigation, and object-directed navigation into a unified interface. The agent selects among these modes, inspects execution feedback, and obtains additional information through interaction and memory, connecting action selection with the evidence available for subsequent decisions.

\subsection{Interactive Navigation}
Interactive navigation enables embodied agents to obtain information and assistance through dialogue. CVDN connects navigation with dialogue history collected through interaction with an informed guide~\cite{cvdn2020}. HANNA studies agents that actively request language and visual assistance~\cite{hanna2019}, while ORION integrates thinking, acting, and asking to incorporate feedback into object navigation~\cite{orion2024}. These approaches treat interaction as an action that can influence how an agent interprets its task and proceeds through the environment.

Interaction is particularly useful when the available instruction or observation is insufficient to identify the intended goal. VL-LN and CoIN study dialogue for underspecified goals and target disambiguation~\cite{vlln2025,coin2025}. GG-Nav and ProCompNav further investigate interactive navigation and goal resolution~\cite{ggnav2026,procompnav2026}. Talk2Escape and AVERT-VLN also examine external assistance in navigation~\cite{talk2escape2026,avertvln2026}. Across these settings, useful interaction requires the agent to decide when help is needed, formulate a relevant request, and incorporate the response into subsequent action. Questions therefore connect the agent's current uncertainty with information that supports continued navigation.

\method{} integrates this initiative into the unified navigation loop. The agent can ask for clarification, request guidance, or seek assistance with an environmental difficulty, then use the response and subsequent observations to decide how to proceed. Information acquired through interaction also becomes part of persistent place records, remaining available beyond the current exchange. This connection between asking, acting, and retaining experience is central to {\bfseries\itshape Ask to Evolve}.

\subsection{Spatial Memory and Experience Evolution}
Spatial memory associates observations and semantic information with locations that an embodied agent can revisit. Visual language maps connect semantic queries with spatial locations, while object-centric graphs represent relations among detected entities~\cite{vlmaps2023,conceptgraphs2024}. Three-dimensional scene memory supports selective retrieval of observations for embodied tasks~\cite{threedmem2025}. These representations provide spatial context for interpreting current observations and selecting actions. Their use over time also involves retrieving relevant content and revising stored information when new evidence becomes available.

Lifelong navigation explicitly considers experience reuse across tasks. GOAT-Bench evaluates sequential category, description, and image goals, while Iterative VLN studies continued navigation using retained experience~\cite{goat2024,ivln2023}. Memoir combines viewpoint-level observation and behavior memory with query imagination~\cite{memoir2026}. NavHarness preserves navigation state, maps, and task knowledge across sessions, while HarnessVLN connects planning support with persistent spatial and event representations~\cite{navharness2026,harnessvln2026}. These approaches consider both what an agent remembers and how its decision interface makes that memory available for subsequent navigation.

\method{} focuses on how autonomous action and proactive interaction jointly contribute to persistent spatial experience. Place records connect locations with descriptions, observations, historical images, and dialogue. Information obtained through a question can revise the interpretation of a place, while subsequent movement and observation provide further evidence for updating the record. The agent can inspect historical evidence, revisit remembered locations, and update memory as its understanding changes.
\section{\method{}}
\label{sec:method}


\begin{figure}[!t]
\centering
\includegraphics[width=\linewidth]{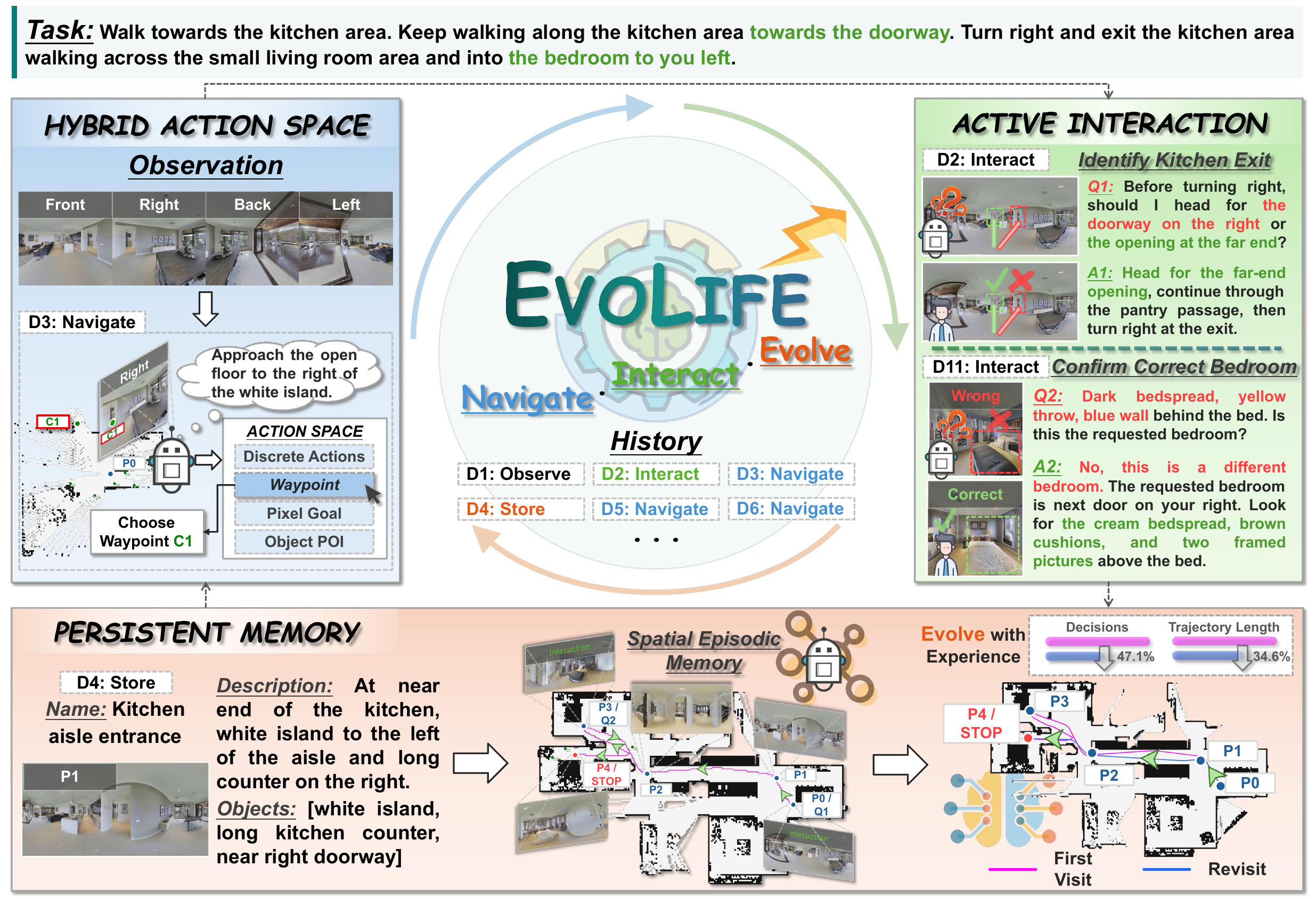}
\caption{Overview of the unified \method{} framework.
A multimodal agent integrates hybrid actions, active interaction, and
persistent memory within a unified decision loop. The agent adaptively
selects navigation actions based on current observations and execution
feedback, actively seeks external guidance to resolve navigation
uncertainties, and retains spatially grounded experience for future tasks.
$C_i$ denotes a geometric candidate waypoint. $P_i$ denotes a
remembered place associated with its recorded pose and supporting evidence.
$D_i$ denotes the $i$-th tool-level decision, encompassing perception,
navigation, interaction, and memory operations.
$Q_i$ and $A_i$ denote the agent-generated question and the corresponding
guide answer in the $i$-th interaction, respectively.
The illustrated example demonstrates how hybrid actions, interaction,
and memory jointly support lifelong embodied navigation.}
\label{fig:overview}
\end{figure}

\subsection{Overview}

We propose \method{}, a unified harness for lifelong embodied navigation that enables an agent to choose how to move, when to ask, and what experience to retain. As illustrated in \figref{fig:overview}, the framework connects three core components: a hybrid action space, active interaction, and persistent memory.

The remainder of this section is organized as follows. Sec.~\ref{sec:problem_formulation} defines the navigation task and tool-based decision process. Sec.~\ref{sec:hybrid_actions} introduces the four navigation modes within the hybrid action space. Sec.~\ref{sec:active_interaction} describes agent-initiated clarification and assistance requests. Sec.~\ref{sec:persistent_memory} presents how spatial experience is recorded, updated, and reused. The principal tools are summarized in \tabref{tab:tools}.

\begin{table*}[t]
\centering
\caption{Principal \method{} tools organized into four functional groups. The agent autonomously selects tools and their parameters.}
\label{tab:tools}

\begingroup
\small
\setlength{\tabcolsep}{5pt}
\renewcommand{\arraystretch}{1.20}
\arrayrulecolor{EvoRule}

\begin{tabularx}{\linewidth}{
@{}
>{\raggedright\arraybackslash\footnotesize}p{.25\linewidth}
>{\raggedright\arraybackslash}p{.24\linewidth}
>{\raggedright\arraybackslash}X
@{}
}
\toprule
\textbf{Tool}
& \textbf{Agent-selected input}
& \textbf{Returned evidence and function} \\
\midrule

\rowcolor{EvoGreyTint}
\multicolumn{3}{l}{\textbf{Perception}} \\

\tool{observe}\newline
\tool{observe\_panorama\_candidates}
& Observation request
& Return current four-view or panoramic observations and geometric waypoint candidates. \\

\tool{observe\_map}
& Memory Map
& Display observed occupancy maps, executed routes, candidate waypoints, and remembered places. \\

\tool{process\_image}
& Source view; processing pipeline and parameters
& Return processed images and their processing records. \\

\tool{localize\_object}\newline
\tool{verify\_object}
& View and box/point prompts; image version or object ID
& Estimate object geometry from RGB-D observations and return visual overlays and depth-consistency evidence. \\

\midrule

\rowcolor{EvoBlueTint}
\multicolumn{3}{l}{\textbf{Hybrid Action Space}} \\

\tool{navigate\_candidate}
& Candidate waypoint index; optional inspection timing
& Navigate to the selected geometric waypoint through observed free space. \\

\tool{navigate\_pixel}
& View; normalized floor pixel coordinates; optional inspection timing
& Project the agent-selected image location into a ground waypoint using RGB-D, verify reachability, and execute navigation. \\

\tool{navigate\_object}
& Object ID; stand-off distance; facing and inspection options
& Plan a path to reachable free space near the localized object and return observations. \\

\tool{step (low-level action)}
& Go forward / Turn left / Turn right
& Execute local discrete actions for exploration or fine adjustment and return observations. \\

\tool{stop}
& Completion summary
& Terminate the current navigation task. \\

\midrule

\rowcolor{EvoOrangeTint}
\multicolumn{3}{l}{\textbf{Active Interaction}} \\

\tool{ask\_human}
& Agent-generated question or assistance request; interaction intent
& Obtain natural-language guidance and retain the dialogue and its context. \\

\midrule

\rowcolor{EvoTealTint}
\multicolumn{3}{l}{\textbf{Persistent Memory}} \\

\tool{remember\_place}
& Place name; description; objects; evidence; optional existing place ID
& Create or update a place record associated with the actual pose, map, timestamp, and saved visual observations. \\

\tool{search\_memory}
& Text query
& Retrieve structured place records by matching stored names, descriptions, and object labels. \\

\tool{inspect\_place}
& Place ID
& Retrieve historical four-view images and place details. \\

\tool{navigate\_place}\newline
\tool{resume\_navigation}
& Place ID or retained destination; optional inspection timing
& Replan a path to a remembered location through currently observed free space. \\

\bottomrule
\end{tabularx}
\endgroup
\end{table*}

\subsection{Problem Formulation}
\label{sec:problem_formulation}

We consider embodied navigation tasks in previously unseen environments, where the agent is not provided with a pre-built map. Instead, it must navigate through online observations and execution feedback, requiring strong generalization to unseen scenes and robustness to partial observability, spatial uncertainty, and ambiguous goals.

Let $g_j$ denote the active navigation goal at task stage $j$, $x_t$ the environment state at decision step $t$, and $o_t$ the latest current observations. Depending on the task, $g_j$ can contain a language instruction, an object category, a target description, or a reference image. The ordered history of tool requests and returns during the current execution is $H_t=((a_\tau,z_\tau))_{\tau=0}^{t-1}$, with $H_0$ empty. Persistent memory is represented as $M_t=(\mathcal{G}_t,\mathcal{P}_t,\mathcal{R}_t,\mathcal{D}_t)$, where $\mathcal{G}_t$ stores observed geometry, $\mathcal{P}_t$ place records, $\mathcal{R}_t$ executed route history, and $\mathcal{D}_t$ retained dialogue.

At each decision step, the agent selects a tool request through the multimodal decision process:
\begin{equation}
a_t=(u_t,\eta_t) =\pi_\theta(g_j,o_t,H_t,M_t),
\label{eq:policy}
\end{equation}
where $u_t$ is the selected tool and $\eta_t$ its arguments. Here, $M_t$ denotes the persistent memory available to the agent. The request is executed in the environment, which returns observations and feedback:
\begin{equation}
\begin{aligned}
(x_{t+1},o_{t+1},z_t) &=\mathcal{E}(x_t,M_t,a_t),\\
M_{t+1} &=\mathcal{U}(M_t,a_t,z_t,o_{t+1}),\\
H_{t+1} &=H_t\mathbin{\|}\bigl[(a_t,z_t)\bigr],
\end{aligned}
\label{eq:update}
\end{equation}
where $\mathcal{E}$ denotes tool execution, $\mathcal{U}$ memory updating, and $\|$ sequence concatenation. The feedback $z_t$ records the actual tool outcome. Calling \tool{stop} submits the current task, after which the environment terminates the task or advances to the next task stage.

\subsection{Hybrid Action Space}
\label{sec:hybrid_actions}

The movement interface supports four complementary navigation modes:
\begin{equation}
\mathcal{A}^{\mathrm{move}}_t = \mathcal{A}^{\mathrm{wp}}_t \cup \mathcal{A}^{\mathrm{local}} \cup \mathcal{A}^{\mathrm{pixel}}_t \cup \mathcal{A}^{\mathrm{obj}}_t,
\label{eq:actions}
\end{equation}
where the four sets contain requests for waypoint navigation, local discrete adjustments, pixel-based navigation, and object-directed navigation, respectively. These movement requests remain available alongside perception, interaction, memory, and stopping tools in Eq.~\eqref{eq:policy}. Perception tools provide the evidence used to select movement. The \tool{observe} tool returns current four-view observations, while \tool{observe\_panorama\_candidates} provides a panoramic view annotated with geometric waypoint proposals~\cite{clash2025}. The \tool{observe\_map} tool visualizes observed occupancy, available floor layers, executed routes, candidates, and remembered places. The \tool{process\_image} tool returns processed visual evidence together with its processing record.

For \textbf{waypoint navigation}, let $C_t$ denote geometric candidates generated from observed geometry and refreshed using the latest RGB-D observations. The agent assesses these candidates using the available views and selects an index for \tool{navigate\_candidate}. The backend then plans through observed free space and executes the resulting path continuously. Navigation to a remembered place uses the same waypoint execution mechanism, as described in Sec.~\ref{sec:persistent_memory}. 

\textbf{Local discrete adjustments} are provided by \tool{step}, which supports forward movement and left or right turns. In the simulated interface, \tool{step(1)} requests a forward movement of $0.25$\,m, while \tool{step(2)} and \tool{step(3)} turn left and right by $15^\circ$, respectively. These actions support local exploration, orientation changes, and fine adjustment after a longer navigation request, with observations and feedback returned for the next decision.

\textbf{Pixel-based navigation} allows the agent to select a floor location directly in an image through \tool{navigate\_pixel}. Its arguments identify the source view, normalized pixel coordinates, and image version. The backend uses RGB-D evidence and the corresponding camera transform to obtain a ground waypoint, checks its reachability through observed free space, and executes navigation toward it. This mode connects a visually selected location with continuous movement without requiring selection from the candidate set $C_t$.


For \textbf{object-directed navigation}, the agent first identifies the
target in the current RGB observations and uses
\tool{localize\_object} to select a source view and specify a bounding
box or image point for RGB-D localization.
The backend filters valid depth measurements within the selected box
or a local neighborhood of the selected point, extracts a
depth-consistent component, and back-projects the retained pixels to
estimate object geometry.
The returned object identifier links this estimate to its supporting
3D points and visual overlays.
The agent can use \tool{verify\_object} to reproject the estimate into
current views and inspect depth-consistency evidence.
The \tool{navigate\_object} tool then accepts the object identifier,
a stand-off distance, and optional facing and inspection settings.
The backend searches for reachable approach locations in observed
free space and plans a path to a selected location.
The agent examines the returned views and execution feedback to
assess the approach and decide whether further movement or adjustment
is needed.
The agent remains responsible for identifying the intended target
from visual evidence and deciding when to stop.

Navigation requests can specify an inspection distance or execution time. Control returns at the requested checkpoint, or upon arrival, collision, lack of progress, or budget exhaustion. The agent then selects its next tool using the actual returned state. Navigation requests perform movement, while task submission remains a \tool{stop} decision.

\subsection{Active Interaction}
\label{sec:active_interaction}

The agent uses \tool{ask\_human} to request clarification, guidance, or assistance. It autonomously determines when to interact and what to ask based on the active goal, current observations, and relevant experience. When $u_t=\texttt{ask\_human}$, the arguments $\eta_t$ specify the question or assistance request $q_t$ and its intent.

In simulation, we employ a separate multimodal model as an online guide, whose context is independent of the navigation agent. Upon receiving an interaction request, the guide generates a natural-language response:
\begin{equation}
r_t = \mathcal{O}_{\psi}
\bigl(q_t,H_t^{\mathrm{dlg}},\Xi_t\bigr),
\label{eq:oracle}
\end{equation}
where $\mathcal{O}_{\psi}$ denotes the guide model, $H_t^{\mathrm{dlg}}$ represents the relevant historical question--answer exchanges, and $\Xi_t$ contains guide-side task and scene evidence.


\begin{figure}[!t]
\centering
\includegraphics[width=\linewidth]{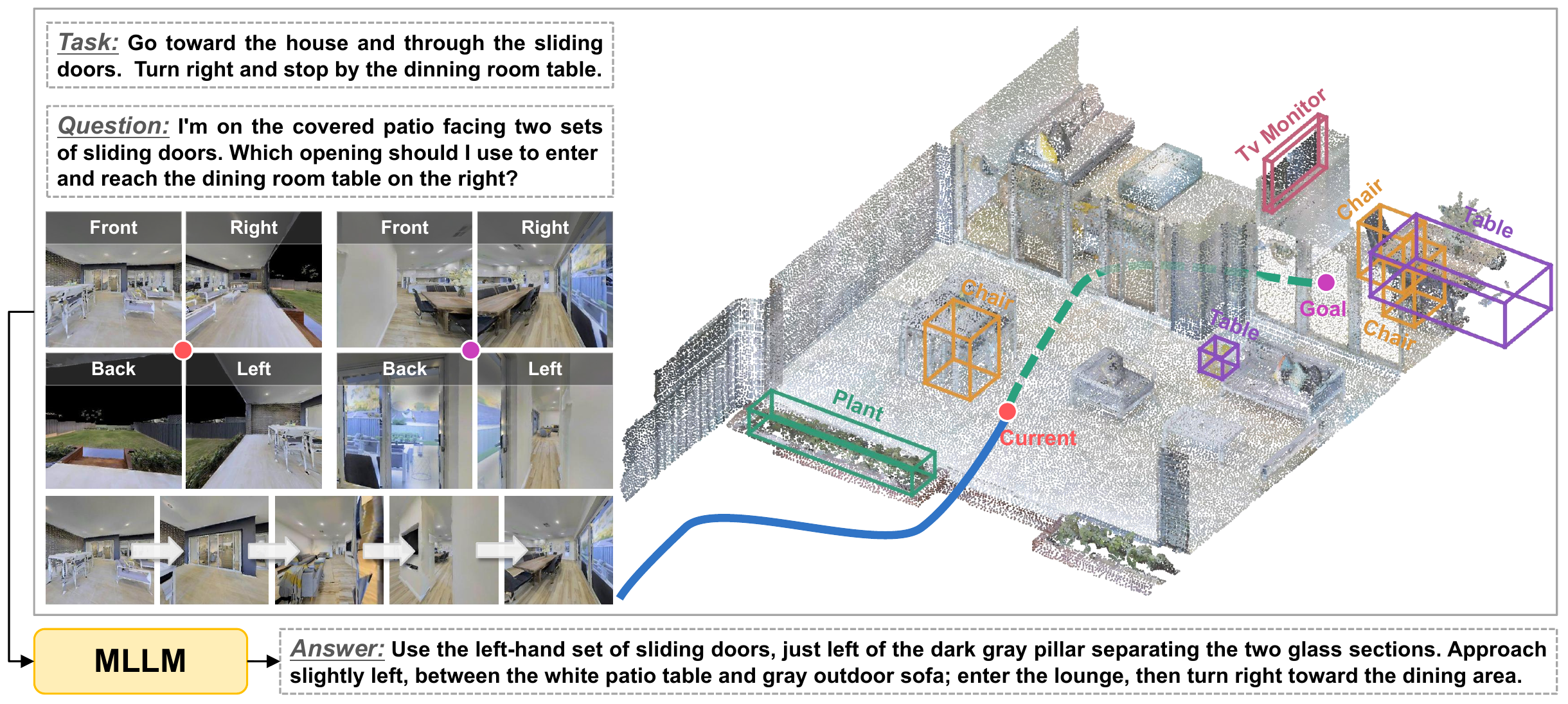}
\caption{Online guide response generation. The navigation agent actively generates a question, while an independent multimodal guide generates a natural-language response based on the navigation instruction, current observations, target information, and reference route. The figure illustrates the current and target views, reference route, and spatial relationships between the agent and the destination. Only the generated natural-language response is returned to the navigation agent, which independently makes subsequent decisions. This process simulates natural human-to-human interaction, providing navigation guidance through directions, landmarks, and spatial relationships.}
\label{fig:oracle}
\end{figure}

As illustrated in \figref{fig:oracle}, the guide receives the navigation instruction, current visual observations, target information, and historical dialogue. Unlike the navigator, the guide additionally has access to privileged spatial information, including the current pose, target location, shortest navigable path, and relevant map information. Visual representations of the reference route are also provided to facilitate spatial reasoning and natural-language guidance.

Rather than directly exposing target coordinates, complete reference paths, or private maps to the navigator, the guide translates this information into natural-language responses using relative directions, visible landmarks, and spatial relationships, resembling how humans provide directions in unfamiliar environments. For instance-level navigation tasks such as VL-LN and CoIN, the navigator initially receives only the target category, whereas the guide has access to the specific target instance, allowing it to resolve instance ambiguity through interaction.

The generated response $r_t$ is returned through the tool feedback $z_t$ and becomes available for subsequent decisions. The agent can use the guidance to clarify the intended target, determine a navigation direction, or resolve difficulties encountered during execution. It then selects its next action and verifies the outcome through new observations and execution feedback. In physical deployment, interaction may also involve direct human assistance, such as opening a closed door, after which the agent observes the changed environment and resumes navigation.

Interaction also contributes to persistent experience. The question, response, and associated spatial context can be retained as place-grounded memory, allowing acquired guidance to be retrieved and reused during subsequent navigation tasks. In this way, active interaction supports both immediate navigation decisions and the accumulation of experience for future encounters.

\subsection{Persistent Memory}
\label{sec:persistent_memory}

Persistent memory enables spatial experience to be retained across task boundaries and reused during subsequent navigation tasks. Rather than resetting acquired knowledge after each task, the agent preserves observations, executed motion, and dialogue associated with spatial locations, allowing previous encounters to inform future decisions. Observed geometry $\mathcal{G}_t$ supports planning, executed routes $\mathcal{R}_t$ preserve movement history, and retained dialogue $\mathcal{D}_t$ provides interaction evidence. The place collection $\mathcal{P}_t$ organizes this accumulated experience around named locations that the agent can inspect, revise, and revisit.

A place record $m_i\in\mathcal{P}_t$ is represented as
\begin{equation}
m_i= \bigl( \mathrm{id}_i,p_i,d_i,O_i,I_i,D_i,c_i,\tau_i,\sigma_i \bigr),
\label{eq:place}
\end{equation}
where $\mathrm{id}_i$ is the place identifier, $p_i$ the recorded pose, $d_i$ its name and description, $O_i$ the object labels, $I_i$ the historical views, and $D_i$ the associated dialogue evidence linked to $\mathcal{D}_t$. The context $c_i$ identifies the associated map, $\tau_i$ records the storage time, and $\sigma_i$ indicates whether the record is active or retired. These fields summarize linked records grounded in actual poses, observations, and interactions.

The \tool{remember\_place} tool creates a record at the actual current pose or updates an existing place identifier, preserving timestamped visual evidence alongside its description. Place descriptions may already accompany current observations. When further retrieval is useful, \tool{search\_memory} matches stored names, descriptions, and object labels; an empty query lists available places. The \tool{inspect\_place} tool retrieves historical four-view images and place details, explicitly distinguished from current observations. The agent selects these operations according to its information needs, without a fixed retrieval order.

Given an available place identifier, \tool{navigate\_place} plans toward its recorded location using current observed geometry. The destination can remain active during local inspection or avoidance, after which \tool{resume\_navigation} replans from the actual current pose. Both tools use the waypoint execution mechanism, allowing current observations and motion feedback to guide continued navigation.

Geometry, executed routes, and spatially associated dialogue are recorded automatically, while semantic place records are maintained selectively by the agent. New observations and answers can enrich a description, correct a mistaken interpretation, or motivate retiring an obsolete record. For example, a clarification can establish which room an entrance leads to, while a later visit can provide additional visual landmarks or reveal changed access conditions. Crucially, these records remain available beyond individual tasks, allowing the agent to retrieve and reuse spatial knowledge accumulated during earlier encounters. During revisits, retained descriptions, historical images, and dialogue help distinguish similar locations and guide navigation toward remembered places. New observations and execution feedback can then verify, refine, or correct existing records. Through this continuous process of retention, reuse, and revision, spatial experience progressively evolves across tasks, enabling future navigation to build on earlier exploration and interaction rather than repeatedly acquiring the same information.

\section{Experiments}
\label{sec:experiments}

\subsection{Experimental Setup and Protocols}
\label{sec:setup}

\paragraph{Tasks and Benchmarks.}
\tabref{tab:benchmark_protocols} summarizes the evaluation protocols. R2R-CE and RxR-CE evaluate route following in continuous environments~\cite{r2r2018,rxr2020,vlnce2020}. HM3D v2 and HM3D-OVON use object-category goals in indoor environments~\cite{hm3d2021,hm3dsem2023,ovon2024}. VL-LN and CoIN start from category-only, underspecified language goals and use interaction to identify the intended target instance~\cite{vlln2025,coin2025}. LHPR-VLN provides ordered multi-stage tasks, while GOAT-Bench combines object-category, text-description, and reference-image goals~\cite{lhpr2025,goat2024}.
We evaluate 100 episodes for each of the six single-task benchmarks, 50 LHPR sequences containing 133 stages, and 50 GOAT sequences containing 372 subtasks. The main evaluation thus comprises 700 complete episode and 1,105 subtasks.
R2R-CE reuses the fixed 100-episode evaluation set from OpenNav. RxR-CE, HM3D v2, HM3D-OVON, VL-LN, and CoIN each use 100 randomly selected episodes.
After checking dataset annotation consistency, the eligible pools contain 175 of 205 LHPR-VLN sequences and 354 of 360 GOAT sequences. We randomly select 50 complete sequences from each eligible pool for evaluation. Within each sequence, memory persists across subtasks, which are executed continuously. Memory is reset at the beginning of each new sequence.

\begin{table*}[t]
\centering
\arrayrulecolor{EvoRule}
\centering
\caption{Evaluation protocols and sample sizes across eight navigation
benchmarks. An episode denotes one complete task or continuous sequence.}
\label{tab:benchmark_protocols}
\begingroup
\footnotesize
\setlength{\tabcolsep}{4pt}
\renewcommand{\arraystretch}{1.18}

\begin{tabularx}{\linewidth}{
@{}l
>{\raggedright\arraybackslash}X
cc
>{\raggedright\arraybackslash}X
>{\raggedright\arraybackslash}p{0.14\linewidth}
@{}
}
\toprule
Benchmark
& Task input
& Episodes
& Subtasks
& Success criterion
& Metrics \\
\midrule

R2R-CE
& Route instruction
& 100 & --
& STOP within $3\,\mathrm{m}$ of the destination
& SR, OSR, SPL,\newline NE \\

RxR-CE
& English route instruction
& 100 & --
& STOP within $3\,\mathrm{m}$ of the destination
& SR, SPL,\newline NE, nDTW \\

HM3D-OVON
& Open-vocabulary object category; Val-Unseen
& 100 & --
& STOP with $d_g < 0.25\,\mathrm{m}$ to an accepted goal viewpoint
& SR, SPL, NE \\

HM3D v2
& Object category; validation split
& 100 & --
& STOP with $d_g < 0.1\,\mathrm{m}$ to an accepted goal viewpoint
& SR, SPL, NE \\

VL-LN
& Category-only underspecified language goal; IIGN
& 100 & --
& STOP with $d_g < 0.25\,\mathrm{m}$ to an accepted target viewpoint
& \shortstack[l]{SR, SPL, NQ} \\

CoIN
& Category-only language goal; hidden target instance; Unseen categories
& 100 & --
& STOP with $d_g < 0.25\,\mathrm{m}$ to an accepted target-instance viewpoint
& \shortstack[l]{SR, SPL, NQ} \\

\midrule

LHPR-VLN
& One complete instruction containing ordered navigation stages
& 50 & 133
& Each stage: STOP with $d_g < 1\,\mathrm{m}$ to its selected target point;
all stages must succeed for episode success
& SR, ISR, CSR,\newline CGT \\

GOAT-Bench
& Sequential category, description, or image goals; Val-Unseen
& 50 & 372
& Each subtask: STOP with $d_g < 0.25\,\mathrm{m}$ to an accepted goal viewpoint;
all subtasks must succeed for episode success
& eSR, SR, SPL \\

\bottomrule
\end{tabularx}

\endgroup
\end{table*}


\paragraph{Experimental Details.}
Simulation and local tool execution use a single NVIDIA RTX PRO 6000
Blackwell GPU.
The simulation environment is built upon Habitat~\cite{habitat2019}.
The navigator receives four RGB views, depth-derived observed geometry,
and the current pose provided by the simulator.
For the main experiments, both the navigator and the guide use
GPT-6 Astra~\cite{openai2026gpt6astra} with the \texttt{ultra} reasoning setting.
They operate in independent sessions with separate contexts.
The guide privately accesses route- and goal-related information,
while the navigator receives only the natural-language replies
generated by the guide, with a maximum of ten interactions per subtask.
We additionally evaluate Qwen3.8-Max (\texttt{xhigh})~\cite{qwen2026qwen38max}
and DeepSeek V4.1-Flash (\texttt{max})~\cite{deepseek2026v41flash}.
In each configuration, the navigator and guide use the same backbone
in separate sessions, while retaining the shared navigation harness
and tool interfaces.

\paragraph{Measures.}
Success rate (SR) requires the agent to stop within the task's accepted success region. Oracle success rate (OSR) measures whether the navigation trajectory ever reaches the success region, while navigation error (NE) measures the final distance to the goal. Success weighted by path length (SPL) jointly evaluates task success and travel efficiency~\cite{evaluation2018}. Normalized dynamic time warping (nDTW) further evaluates the correspondence between the executed trajectory and the reference path~\cite{ndtw2019}. Trajectory length (TL) measures the total distance traveled by the agent along its executed trajectory, expressed in meters.
For VL-LN and CoIN, we additionally report the Number of Questions
(NQ), defined as the average number of agent-initiated
question--answer exchanges per episode.
For LHPR-VLN, we retain the official evaluation
metrics~\cite{lhpr2025}.
Success Rate (SR) measures whether all subtasks are completed
successfully in the required order, while Independent Success Rate
(ISR) measures individual subtask success.
Conditional Success Rate (CSR) accounts for dependencies between
consecutive subtasks, assigning greater credit when both the current
and preceding subtasks succeed.
CSR weighted by Ground Truth (CGT) further weights these contributions
by the relative ground-truth path lengths of the subtasks.
GOAT distinguishes subtask success rate (s-SR) from complete-sequence success rate (e-SR).
R2R and RxR use a 3\,m destination threshold. For success criteria based on geodesic distance to accepted target viewpoints, the thresholds are 0.25\,m for OVON, VL-LN, CoIN, and GOAT, and 0.1\,m for HM3D v2. LHPR stages use a 1\,m threshold.

\subsection{Comparisons on Eight Benchmarks}
\begin{table*}[!tp]
\centering
\arrayrulecolor{EvoRule}
\caption{Navigation performance on R2R-CE, RxR-CE, HM3D v2, and HM3D-OVON Val-Unseen.}
\label{tab:main_benchmarks}
\begingroup
\fontsize{8.2}{10.0}\selectfont
\setlength{\tabcolsep}{2.5pt}
\renewcommand{\arraystretch}{1.12}
\begin{adjustbox}{max width=\linewidth}
\begin{tabular}{@{}>{\raggedright\arraybackslash}m{43mm}
 cccc@{\hspace{9pt}}cccc@{\hspace{9pt}}ccc@{\hspace{9pt}}ccc@{}}
\toprule
\multirow{2}{*}{\textbf{Method}}
& \multicolumn{4}{c}{\textbf{R2R-CE}}
& \multicolumn{4}{c}{\textbf{RxR-CE}}
& \multicolumn{3}{c}{\textbf{HM3D v2}}
& \multicolumn{3}{c}{\textbf{HM3D-OVON}} \\
\cmidrule(lr){2-5}\cmidrule(lr){6-9}\cmidrule(lr){10-12}\cmidrule(l){13-15}
& NE$\downarrow$
& OSR$\uparrow$
& SR$\uparrow$
& SPL$\uparrow$
& NE$\downarrow$
& nDTW$\uparrow$
& SR$\uparrow$
& SPL$\uparrow$
& NE$\downarrow$
& SR$\uparrow$
& SPL$\uparrow$
& NE$\downarrow$
& SR$\uparrow$
& SPL$\uparrow$ \\
\midrule
NavJev~\cite{navjev2026} & 7.48 & 35.0 & 27.0 & 22.4 & -- & -- & -- & -- & -- & -- & -- & -- & -- & -- \\
Seq2Seq~\cite{vlnce2020} & 7.37 & 40.0 & 32.0 & 30.0 & -- & -- & -- & -- & -- & -- & -- & -- & -- & -- \\
NaVid~\cite{navid2024} & 5.47 & 49.1 & 37.4 & 35.9 & 8.41 & \multicolumn{1}{c}{--} & 23.8 & 21.2 & \multicolumn{1}{c}{--} & \multicolumn{1}{c}{--} & \multicolumn{1}{c}{--} & \multicolumn{1}{c}{--} & \multicolumn{1}{c}{--} & \multicolumn{1}{c}{--} \\
MLANet~\cite{mlanet2023} & 6.30 & 42.0 & 38.0 & 35.0 & -- & -- & -- & -- & -- & -- & -- & -- & -- & -- \\
Uni-NaVid~\cite{uninavid2024} & 5.58 & 53.3 & 47.0 & 42.7 & 6.24 & \multicolumn{1}{c}{--} & 48.7 & 40.9 & \multicolumn{1}{c}{--} & \multicolumn{1}{c}{--} & \multicolumn{1}{c}{--} & \multicolumn{1}{c}{--} & 39.5 & 19.8 \\
P2DNav~\cite{p2dnav2026} & 6.61 & 65.0 & 50.0 & 30.6 & 8.55 & 33.7 & 35.0 & 19.2 & -- & -- & -- & -- & -- & -- \\
ActiveVLN~\cite{activevln2025} & 5.31 & 58.3 & 50.1 & 43.7 & 5.84 & 58.1 & 50.7 & 41.2 & \multicolumn{1}{c}{--} & \multicolumn{1}{c}{--} & \multicolumn{1}{c}{--} & \multicolumn{1}{c}{--} & \multicolumn{1}{c}{--} & \multicolumn{1}{c}{--} \\
NaVILA~\cite{navila2025} & 5.22 & 62.5 & 54.0 & 49.0 & 6.77 & 58.8 & 49.3 & 44.0 & \multicolumn{1}{c}{--} & \multicolumn{1}{c}{--} & \multicolumn{1}{c}{--} & \multicolumn{1}{c}{--} & \multicolumn{1}{c}{--} & \multicolumn{1}{c}{--} \\
DecoVLN~\cite{decovln2026} & 5.01 & 63.5 & 56.3 & 50.5 & 5.73 & 63.5 & 54.2 & 46.3 & \multicolumn{1}{c}{--} & \multicolumn{1}{c}{--} & \multicolumn{1}{c}{--} & \multicolumn{1}{c}{--} & \multicolumn{1}{c}{--} & \multicolumn{1}{c}{--} \\
StreamVLN~\cite{streamvln2025} & 4.98 & 64.2 & 56.9 & 51.9 & 6.22 & 61.9 & 52.9 & 46.0 & \multicolumn{1}{c}{--} & \multicolumn{1}{c}{--} & \multicolumn{1}{c}{--} & \multicolumn{1}{c}{--} & \multicolumn{1}{c}{--} & \multicolumn{1}{c}{--} \\
ETPNav~\cite{etpnav2025} & 4.71 & 65.0 & 57.0 & 49.0 & 5.64 & 61.9 & 54.8 & 44.9 & -- & -- & -- & -- & -- & -- \\
Qwen-VLA-Instruct~\cite{qwenvla2026} & 5.10 & 69.0 & 57.5 & 51.2 & 5.80 & 57.1 & 59.6 & 47.8 & \multicolumn{1}{c}{--} & \multicolumn{1}{c}{--} & \multicolumn{1}{c}{--} & \multicolumn{1}{c}{--} & \multicolumn{1}{c}{--} & \multicolumn{1}{c}{--} \\
InternVLA-N1~\cite{internvlan12025} & 4.83 & 63.3 & 58.2 & 54.0 & 5.91 & 65.3 & 53.5 & 46.1 & \multicolumn{1}{c}{--} & \multicolumn{1}{c}{--} & \multicolumn{1}{c}{--} & \multicolumn{1}{c}{--} & \multicolumn{1}{c}{--} & \multicolumn{1}{c}{--} \\
RynnBrain-Nav-8B~\cite{rynnbrain2026} & 4.92 & 71.6 & 58.6 & 49.6 & 6.20 & 59.6 & 56.1 & 49.6 & \multicolumn{1}{c}{--} & \multicolumn{1}{c}{--} & \multicolumn{1}{c}{--} & \multicolumn{1}{c}{--} & \multicolumn{1}{c}{--} & \multicolumn{1}{c}{--} \\
JanusVLN~\cite{janusvln2025} & 4.78 & 65.2 & 60.5 & 56.8 & 6.06 & 62.1 & 56.2 & 47.5 & \multicolumn{1}{c}{--} & \multicolumn{1}{c}{--} & \multicolumn{1}{c}{--} & \multicolumn{1}{c}{--} & \multicolumn{1}{c}{--} & \multicolumn{1}{c}{--} \\
DualVLN~\cite{dualvln2025} & 4.05 & 70.7 & 64.3 & 58.5 & 4.58 & 70.0 & 61.4 & 51.8 & \multicolumn{1}{c}{--} & \multicolumn{1}{c}{--} & \multicolumn{1}{c}{--} & \multicolumn{1}{c}{--} & \multicolumn{1}{c}{--} & \multicolumn{1}{c}{--} \\
CLASH~\cite{clash2025} & 4.06 & 73.0 & 65.0 & 55.0 & -- & -- & -- & -- & -- & -- & -- & -- & -- & -- \\
Talk2Escape~\cite{talk2escape2026} & 4.80 & 72.0 & 66.0 & 49.4 & 5.89 & 49.9 & 62.9 & 34.2 & \multicolumn{1}{c}{--} & \multicolumn{1}{c}{--} & \multicolumn{1}{c}{--} & \multicolumn{1}{c}{--} & \multicolumn{1}{c}{--} & \multicolumn{1}{c}{--} \\
AstraNav-World~\cite{astranavworld2025} & 3.86 & 73.9 & 67.9 & 65.4 & 3.82 & \multicolumn{1}{c}{--} & 72.9 & 61.5 & \multicolumn{1}{c}{--} & \multicolumn{1}{c}{--} & \multicolumn{1}{c}{--} & \multicolumn{1}{c}{--} & \multicolumn{1}{c}{--} & \multicolumn{1}{c}{--} \\
LightNav-0~\cite{wang2026lightnav}  & 3.91 & 73.7 & 68.5 & 62.8 & 3.66 & 67.4 & 73.6 & 64.5 & \multicolumn{1}{c}{--} & 77.2 & 41.5 & \multicolumn{1}{c}{--} & 47.0 & 24.2 \\
ASENA-VLN-4B~\cite{asena2026} & 3.52 & 72.6 & 68.7 & 64.2 & 3.90 & 73.1 & 70.2 & 59.7 & \multicolumn{1}{c}{--} & \multicolumn{1}{c}{--} & \multicolumn{1}{c}{--} & \multicolumn{1}{c}{--} & \multicolumn{1}{c}{--} & \multicolumn{1}{c}{--} \\
OmniNav~\cite{omninav2025} & 3.74 & 74.6 & 69.5 & 66.1 & 3.77 & \multicolumn{1}{c}{--} & 73.6 & 62.0 & \multicolumn{1}{c}{--} & \multicolumn{1}{c}{--} & \multicolumn{1}{c}{--} & \multicolumn{1}{c}{--} & \multicolumn{1}{c}{--} & \multicolumn{1}{c}{--} \\
Qwen-RobotNav-8B~\cite{qwenrobotnav2026} & 3.53 & 78.5 & 72.1 & 66.6 & 3.58 & 72.5 & 76.5 & 65.7 & \multicolumn{1}{c}{--} & 71.2 & 33.0 & \multicolumn{1}{c}{--} & 51.2 & 24.0 \\
AVERT-VLN~\cite{avertvln2026} & 2.60 & 82.5 & 76.2 & 63.4 & 3.83 & 72.0 & 66.3 & 48.0 & \multicolumn{1}{c}{--} & \multicolumn{1}{c}{--} & \multicolumn{1}{c}{--} & \multicolumn{1}{c}{--} & \multicolumn{1}{c}{--} & \multicolumn{1}{c}{--} \\
NavGPT VLA-8B~\cite{navgpt32026} & 2.97 & 82.5 & 76.9 & 70.6 & 2.65 & 76.7 & 82.4 & 70.6 & \multicolumn{1}{c}{--} & \multicolumn{1}{c}{--} & \multicolumn{1}{c}{--} & \multicolumn{1}{c}{--} & \multicolumn{1}{c}{--} & \multicolumn{1}{c}{--} \\
Robostral Navigate-8B~\cite{robostral2026} & 3.20 & 81.3 & 77.4 & 74.2 & 3.47 & \multicolumn{1}{c}{--} & 75.1 & 68.7 & \multicolumn{1}{c}{--} & \multicolumn{1}{c}{--} & \multicolumn{1}{c}{--} & \multicolumn{1}{c}{--} & \multicolumn{1}{c}{--} & \multicolumn{1}{c}{--} \\
NavGPT-3~\cite{navgpt32026} & 2.18 & 87.0 & 81.5 & 70.4 & 1.56 & 78.5 & 90.4 & 74.3 & \multicolumn{1}{c}{--} & \multicolumn{1}{c}{--} & \multicolumn{1}{c}{--} & \multicolumn{1}{c}{--} & \multicolumn{1}{c}{--} & \multicolumn{1}{c}{--} \\
VLFM~\cite{vlfm2024} & \multicolumn{1}{c}{--} & \multicolumn{1}{c}{--} & \multicolumn{1}{c}{--} & \multicolumn{1}{c}{--} & \multicolumn{1}{c}{--} & \multicolumn{1}{c}{--} & \multicolumn{1}{c}{--} & \multicolumn{1}{c}{--} & \multicolumn{1}{c}{--} & 63.6 & 32.5 & \multicolumn{1}{c}{--} & 35.2 & 19.6 \\
SG-Nav~\cite{sgnav2024} & \multicolumn{1}{c}{--} & \multicolumn{1}{c}{--} & \multicolumn{1}{c}{--} & \multicolumn{1}{c}{--} & \multicolumn{1}{c}{--} & \multicolumn{1}{c}{--} & \multicolumn{1}{c}{--} & \multicolumn{1}{c}{--} & \multicolumn{1}{c}{--} & 49.6 & 25.5 & \multicolumn{1}{c}{--} & \multicolumn{1}{c}{--} & \multicolumn{1}{c}{--} \\
MTU3D~\cite{mtu3d2025} & \multicolumn{1}{c}{--} & \multicolumn{1}{c}{--} & \multicolumn{1}{c}{--} & \multicolumn{1}{c}{--} & \multicolumn{1}{c}{--} & \multicolumn{1}{c}{--} & \multicolumn{1}{c}{--} & \multicolumn{1}{c}{--} & \multicolumn{1}{c}{--} & \multicolumn{1}{c}{--} & \multicolumn{1}{c}{--} & \multicolumn{1}{c}{--} & 40.8 & 12.1 \\
NavFoM~\cite{navfom2025} & 4.61 & 72.1 & 61.7 & 55.3 & 4.74 & 65.8 & 64.4 & 56.2 & \multicolumn{1}{c}{--} & \multicolumn{1}{c}{--} & \multicolumn{1}{c}{--} & \multicolumn{1}{c}{--} & 45.2 & 31.9 \\
ABot-N0~\cite{abotn02026} & 3.78 & 70.8 & 66.4 & 63.9 & 3.83 & \multicolumn{1}{c}{--} & 69.3 & 60.0 & \multicolumn{1}{c}{--} & \multicolumn{1}{c}{--} & \multicolumn{1}{c}{--} & \multicolumn{1}{c}{--} & 54.0 & 30.5 \\
ApexNav~\cite{apexnav2025} & \multicolumn{1}{c}{--} & \multicolumn{1}{c}{--} & \multicolumn{1}{c}{--} & \multicolumn{1}{c}{--} & \multicolumn{1}{c}{--} & \multicolumn{1}{c}{--} & \multicolumn{1}{c}{--} & \multicolumn{1}{c}{--} & 1.31 & 76.2 & 38.0 & \multicolumn{1}{c}{--} & \multicolumn{1}{c}{--} & \multicolumn{1}{c}{--} \\
ASENA~\cite{asena2026} & \multicolumn{1}{c}{1.84} & \multicolumn{1}{c}{90.0} & \multicolumn{1}{c}{87.0} & 72.4 & \multicolumn{1}{c}{0.82} & \multicolumn{1}{c}{--} & \multicolumn{1}{c}{93.0} & 78.0 & \multicolumn{1}{c}{--} & 80.8 & 38.1 & \multicolumn{1}{c}{--} & 64.9 & 38.4 \\
SuperNav~\cite{supernav2026}
& -- & -- & -- & --
& -- & -- & -- & --
& -- & 80.3 & 33.4
& -- & 68.3 & 38.4 \\
AECNav~\cite{aecnav2026}
& -- & -- & -- & --
& -- & -- & -- & --
& -- & 84.7 & 45.3
& -- & 57.3 & 30.5 \\
\midrule
\rowcolor{EvoTealTint}
\textbf{\method{} (Ours)} & \textbf{0.95} & \textbf{100} & \textbf{100} & \textbf{82.2} & \textbf{0.71} & \textbf{82.3} & \textbf{100} & \textbf{75.4} & \textbf{0.04} & \textbf{100} & \textbf{64.3} & \textbf{0.10} & \textbf{100} & \textbf{71.4} \\
\bottomrule
\end{tabular}
\end{adjustbox}
\endgroup
\end{table*}
\begin{table*}[t]
\centering
\arrayrulecolor{EvoRule}
\centering
\caption{Comparison on VL-LN, CoIN, LHPR-VLN, and GOAT-Bench.}
\label{tab:four_benchmarks}
\begingroup
\scriptsize
\setlength{\tabcolsep}{2.5pt}
\renewcommand{\arraystretch}{1.10}
\resizebox{\linewidth}{!}{%
\begin{tabular}{@{}l
*{3}{r}@{\hspace{8pt}}
*{3}{r}@{\hspace{8pt}}
*{4}{r}@{\hspace{8pt}}
*{3}{r}@{}}
\toprule
\multirow{2}{*}{\textbf{Method}}
& \multicolumn{3}{c}{\textbf{VL-LN}}
& \multicolumn{3}{c}{\textbf{CoIN}}
& \multicolumn{4}{c}{\textbf{LHPR-VLN}}
& \multicolumn{3}{c}{\textbf{GOAT-Bench}} \\
\cmidrule(lr){2-4}
\cmidrule(lr){5-7}
\cmidrule(lr){8-11}
\cmidrule(l){12-14}
& SR$\uparrow$ & SPL$\uparrow$ & NQ$\downarrow$
& SR$\uparrow$ & SPL$\uparrow$ & NQ$\downarrow$
& SR$\uparrow$ & ISR$\uparrow$ & CSR$\uparrow$ & CGT$\uparrow$
& s-SR$\uparrow$ & SPL$\uparrow$ & e-SR$\uparrow$ \\
\midrule
VLFM~\cite{vlfm2024}
& 8.8 & 5.6 & --
& 0.2 & 0.1 & --
& \multicolumn{4}{c}{--}
& \multicolumn{3}{c}{--} \\
MTU3D~\cite{mtu3d2025}
& 12.8 & 5.4 & --
& 8.7 & 3.3 & --
& \multicolumn{4}{c}{--}
& \multicolumn{3}{c}{--} \\
AIUTA~\cite{coin2025}
& 4.0 & 2.2 & 5.42
& 6.7 & 2.3 & 1.13
& \multicolumn{4}{c}{--}
& \multicolumn{3}{c}{--} \\
ProCompNav~\cite{procompnav2026}
& \multicolumn{3}{c}{--}
& 17.0 & 6.2 & 2.30
& \multicolumn{4}{c}{--}
& \multicolumn{3}{c}{--} \\
InternVLA-N1~\cite{internvlan12025,ggnav2026}
& 25.4 & 16.2 & 1.98
& 17.2 & 8.2 & 2.61
& \multicolumn{4}{c}{--}
& \multicolumn{3}{c}{--} \\
GG-Nav~\cite{ggnav2026}
& 42.0 & 23.2 & 3.01
& 19.6 & 8.8 & 1.79
& \multicolumn{4}{c}{--}
& \multicolumn{3}{c}{--} \\

\addlinespace[2pt]
GPT-4 + NaviLLM~\cite{navillm2023,lhpr2025}
& \multicolumn{3}{c}{--}
& \multicolumn{3}{c}{--}
& 0.0 & 2.2 & 1.5 & 2.6
& \multicolumn{3}{c}{--} \\
MGDM~\cite{lhpr2025}
& \multicolumn{3}{c}{--}
& \multicolumn{3}{c}{--}
& 0.0 & 2.3 & 1.7 & 2.9
& \multicolumn{3}{c}{--} \\
Aux-Think~\cite{auxthink2025}
& \multicolumn{3}{c}{--}
& \multicolumn{3}{c}{--}
& 0.7 & 3.2 & 2.0 & 1.5
& \multicolumn{3}{c}{--} \\
FantasyVLN~\cite{fantasyvln2026}
& \multicolumn{3}{c}{--}
& \multicolumn{3}{c}{--}
& 2.4 & 11.0 & 9.6 & 9.0
& \multicolumn{3}{c}{--} \\

\addlinespace[2pt]
MemoryExplorer~\cite{memoryexplorer2026}
& \multicolumn{3}{c}{--}
& \multicolumn{3}{c}{--}
& \multicolumn{4}{c}{--}
& 46.4 & 28.0 & -- \\
SSMG-Nav~\cite{ssmgnav2026}
& \multicolumn{3}{c}{--}
& \multicolumn{3}{c}{--}
& \multicolumn{4}{c}{--}
& 46.5 & 34.1 & 8.6 \\
EvoMemNav~\cite{evomemnav2026}
& \multicolumn{3}{c}{--}
& \multicolumn{3}{c}{--}
& \multicolumn{4}{c}{--}
& 59.6 & 38.9 & -- \\
ReEXplore~\cite{reexplore2025}
& \multicolumn{3}{c}{--}
& \multicolumn{3}{c}{--}
& \multicolumn{4}{c}{--}
& 59.8 & 42.5 & -- \\
AstraNav-Memory~\cite{astranavmemory2025}
& \multicolumn{3}{c}{--}
& \multicolumn{3}{c}{--}
& \multicolumn{4}{c}{--}
& 62.7 & 56.9 & -- \\
STEGNav~\cite{stegnav2026}
& \multicolumn{3}{c}{--}
& \multicolumn{3}{c}{--}
& \multicolumn{4}{c}{--}
& 66.3 & 39.7 & -- \\
GSMem~\cite{gsmem2026}
& \multicolumn{3}{c}{--}
& \multicolumn{3}{c}{--}
& \multicolumn{4}{c}{--}
& 67.2 & 46.9 & -- \\
3D-Mem~\cite{threedmem2025}
& \multicolumn{3}{c}{--}
& \multicolumn{3}{c}{--}
& \multicolumn{4}{c}{--}
& 69.1 & 48.9 & -- \\
MetaNav~\cite{metanav2026}
& \multicolumn{3}{c}{--}
& \multicolumn{3}{c}{--}
& \multicolumn{4}{c}{--}
& 71.4 & 51.8 & -- \\
HGR~\cite{hgr2026}
& \multicolumn{3}{c}{--}
& \multicolumn{3}{c}{--}
& \multicolumn{4}{c}{--}
& 72.4 & 56.2 & -- \\
ObsGraph~\cite{obsgraph2026}
& \multicolumn{3}{c}{--}
& \multicolumn{3}{c}{--}
& \multicolumn{4}{c}{--}
& 72.7 & 51.5 & -- \\
HIMM~\cite{himm2026}
& \multicolumn{3}{c}{--}
& \multicolumn{3}{c}{--}
& \multicolumn{4}{c}{--}
& 72.8 & 56.1 & -- \\
NavHarness~\cite{navharness2026}
& \multicolumn{3}{c}{--}
& \multicolumn{3}{c}{--}
& \multicolumn{4}{c}{--}
& 83.7 & 62.3 & 36.9 \\
RoboAtlas~\cite{roboatlas2026}
& \multicolumn{3}{c}{--}
& \multicolumn{3}{c}{--}
& \multicolumn{4}{c}{--}
& 90.6 & 53.4 & -- \\

\midrule
\rowcolor{EvoTealTint}
\textbf{\method{} (Ours)}
& \textbf{99.0} & \textbf{76.2} & \textbf{3.21}
& \textbf{100} & \textbf{73.5} & \textbf{2.83}
& \textbf{98.0} & \textbf{99.3} & \textbf{98.9} & \textbf{98.8}
& \textbf{99.3} & \textbf{81.9} & \textbf{94.0} \\
\bottomrule
\end{tabular}%
}
\endgroup

\end{table*}

\tabref{tab:main_benchmarks} presents the experimental results on vision-language navigation and object navigation tasks. Our method achieves 100\% SR across all four benchmarks, including R2R-CE, RxR-CE, HM3D v2, and HM3D-OVON, demonstrating excellent navigation performance. On R2R-CE, our method achieves 100\% SR, 82.2\% SPL, and 0.95\,m NE, outperforming ASENA (87.0\% SR) with an absolute SR improvement of 13.0\%. On RxR-CE, our method achieves 100\% SR, 75.4\% SPL, 0.71\,m NE, and 82.3\% nDTW, surpassing ASENA (93.0\% SR) by 7.0\% in SR. For object navigation, our method achieves 100\% SR and 64.3\% SPL on HM3D v2, outperforming AECNav (84.7\% SR) by 15.3\% in SR. On HM3D-OVON, our method achieves 100\% SR and 71.4\% SPL, surpassing SuperNav (68.3\% SR) by 31.7\% in SR.

\tabref{tab:four_benchmarks} further presents the performance of our method on interactive instance navigation and multi-stage continuous navigation tasks. On VL-LN, our method achieves 99.0\% SR and 76.2\% SPL, outperforming GG-Nav~\cite{ggnav2026} (42.0\% SR and 23.22\% SPL) by 57.0\% in SR and 53.0\% in SPL. On CoIN, our method achieves 100\% SR and 73.5\% SPL, substantially outperforming GG-Nav (19.6\% SR and 8.8\% SPL) on the unseen split. These results demonstrate that our method effectively resolves target ambiguity through active interaction, enabling accurate instance localization and navigation.

Our method also achieves excellent performance on more challenging multi-stage navigation tasks. On LHPR-VLN, our method successfully completes 49 out of 50 full task sequences, achieving 98.0\% SR, 99.3\% ISR, 98.9\% CSR, and 98.8\% CGT, substantially outperforming FantasyVLN~\cite{fantasyvln2026} (2.4\% SR). On GOAT-Bench, our method achieves 99.3\% subtask SR, 81.9\% SPL, and 94.0\% episode success rate (e-SR). Compared with NavHarness using the same GPT-6 Astra model (83.7\% s-SR and 36.9\% e-SR), our method achieves improvements of 15.6\% in subtask SR and 57.1\% in e-SR.

These results demonstrate that our method not only effectively completes individual navigation tasks but also supports the continuous execution of complex multi-stage tasks, highlighting the effectiveness of the unified framework integrating hybrid actions, active interaction, and persistent memory.

\subsection{Interaction and Memory Conditions}

\begin{table*}[t]
\centering
\arrayrulecolor{EvoRule}
\caption{Navigation performance and efficiency across different benchmarks
and interaction-memory settings.
In the Interaction column, $\checkmark$ and $\times$ indicate whether
active interaction is enabled or disabled, respectively.
In the Memory column, $\times$ denotes a first-visit run in an unseen
environment, while $\checkmark$ denotes a revisit run that reuses
memory retained from the first visit.
Memory accumulated during the current run remains available in both settings.
ND denotes the mean number of completed tool-level decisions;
NQ denotes the mean number of active interaction exchanges.
For LHPR-VLN and GOAT-Bench, ND and NQ are reported as
complete task / subtask, where one complete task is a continuous sequence.
Subtask averages are computed by dividing the total counts by the total
number of subtasks.
Steps, time, and cost are averaged per episode, or per complete sequence
for the continuous benchmarks.}
\label{tab:interaction_memory}
\small
\setlength{\tabcolsep}{3pt}
\renewcommand{\arraystretch}{1.12}

\resizebox{\linewidth}{!}{%
\begin{tabular}{llccccccccccc}
\toprule
\multirow{2}{*}{\textbf{Dataset}} &
\multirow{2}{*}{\textbf{Model}} &
\multirow{2}{*}{\textbf{Interaction}} &
\multirow{2}{*}{\textbf{Memory}} &
\textbf{OSR}$\uparrow$ &
\textbf{SR}$\uparrow$ &
\textbf{SPL}$\uparrow$ &
\textbf{NE}$\downarrow$ &
\multirow{2}{*}{\textbf{Steps}$\downarrow$} &
\textbf{ND}$\downarrow$ &
\textbf{NQ}$\downarrow$ &
\textbf{Time}$\downarrow$ &
\textbf{Cost}$\downarrow$ \\
& & & &
\textbf{(\%)} &
\textbf{(\%)} &
\textbf{(\%)} &
\textbf{(m)} &
& \textbf{task / sub.} &
\textbf{task / sub.} &
\textbf{(s)} &
\textbf{(\$/ep.)} \\
\midrule

R2R & GPT-6 Astra
& $\times$ & $\times$
& 89 & 86 & 63.7 & 2.86
& 94.06 & 18.71 & 0.00
& 170.17 & 1.22 \\

\rowcolor{EvoOrangeTint}
R2R & GPT-6 Astra
& $\checkmark$ & $\times$
& 100 & 100 & 82.2 & 0.95
& 70.12 & 15.57 & 1.66
& 164.08 & 1.06 \\

\rowcolor{EvoTealTint}
R2R & GPT-6 Astra
& $\times$ & $\checkmark$
& 100 & 100 & 93.3 & 0.95
& 54.43 & 12.31 & 0.00
& 107.41 & 1.11 \\

\addlinespace

R2R & DeepSeek V4.1-Flash
& $\times$ & $\times$
& 68 & 55 & 39.5 & 5.23
& 136.64 & 43.02 & 0.00
& 361.21 & 0.03 \\

\rowcolor{EvoOrangeTint}
R2R & DeepSeek V4.1-Flash
& $\checkmark$ & $\times$
& 99 & 98 & 70.8 & 1.21
& 118.44 & 44.72 & 3.21
& 459.12 & 0.03 \\

\rowcolor{EvoTealTint}
R2R & DeepSeek V4.1-Flash
& $\times$ & $\checkmark$
& 99 & 99 & 86.0 & 1.18
& 69.47 & 18.01 & 0.00
& 141.36 & 0.02 \\

\midrule

RxR & GPT-6 Astra
& $\checkmark$ & $\times$
& 100 & 100 & 75.4 & 0.71
& 107.98 & 24.99 & 1.64
& 190.03 & 1.88 \\

\midrule

HM3D-OVON & GPT-6 Astra
& $\checkmark$ & $\times$
& 100 & 100 & 71.4 & 0.10
& 64.12 & 13.04 & 2.15
& 129.92 & 0.87 \\

HM3D v2 & GPT-6 Astra
& $\checkmark$ & $\times$
& 100 & 100 & 64.3 & 0.04
& 63.55 & 12.50 & 2.02
& 115.49 & 0.83 \\

\midrule

CoIN & GPT-6 Astra
& $\checkmark$ & $\times$
& 100 & 100 & 73.5 & 0.12
& 85.58 & 15.60 & 2.83
& 122.89 & 1.09 \\

VL-LN & GPT-6 Astra
& $\checkmark$ & $\times$
& 100 & 99 & 76.2 & 0.08
& 106.55 & 19.51 & 3.21
& 154.99 & 1.52 \\

\midrule

LHPR-VLN & GPT-6 Astra
& $\checkmark$ & $\times$
& 98 & 98 & -- & 0.58
& 219.74 & 42.40 / 15.94 & 6.82 / 2.56
& 313.78 & 4.63 \\

GOAT-Bench & GPT-6 Astra
& $\checkmark$ & $\times$
& 98 & 94 & 82.2 & 0.15
& 264.90 & 39.98 / 5.37 & 8.78 / 1.18
& 323.64 & 4.33 \\



\bottomrule
\end{tabular}%
}
\end{table*}

\tabref{tab:interaction_memory} compares three experimental conditions on the same 100 R2R tasks. Without active interaction or initial memory, GPT-6 Astra achieves 86\% SR and 63.7\% SPL. Introducing active interaction during first visits improves SR to 100\% and SPL to 82.2\%. During memory-based revisits, the agent loads the place records, historical images, and interaction dialogues retained from the corresponding first visits, while disabling active interaction, achieving 100\% SR and 93.3\% SPL.

Compared with interactive first visits, memory-based revisits reduce the average motion-equivalent steps from 70.12 to 54.43, the number of tool decisions from 15.57 to 12.31, and the navigator-active execution time from 164.08\,s to 107.41\,s. These results demonstrate that the agent can effectively reuse previously accumulated navigation experience, substantially improving navigation efficiency while maintaining task success.

DeepSeek V4.1-Flash exhibits a similar performance trend. Without interaction or initial memory, it achieves 55\% SR and 39.5\% SPL. Introducing active interaction improves SR to 98\% and SPL to 70.8\%. Under the memory-based revisit condition, DeepSeek V4.1-Flash achieves 99\% SR and 86.0\% SPL.

\begin{table*}[t]
\centering
\arrayrulecolor{EvoRule}
\centering
\caption{Backbone comparison of \method{} on R2R-CE.
ND denotes the mean number of completed tool-level decisions;
NQ denotes the mean number of active interaction exchanges.}
\label{tab:backbone}
\small
\setlength{\tabcolsep}{4pt}
\renewcommand{\arraystretch}{1.12}

\resizebox{\linewidth}{!}{%
\begin{tabular}{lrrrrrrrrr}
\toprule
Model
& OSR$\uparrow$ (\%)
& SR$\uparrow$ (\%)
& SPL$\uparrow$ (\%)
& NE$\downarrow$ (m)
& Steps$\downarrow$
& ND$\downarrow$
& NQ$\downarrow$
& Time$\downarrow$ (s)
& Cost$\downarrow$ (\$/ep.) \\
\midrule
\rowcolor{EvoTealTint}
GPT-6 Astra (ultra)
& 100 & 100 & 82.2 & 0.95
& 70.12 & 15.57 & 1.66 & 164.08 & 1.06 \\

Qwen3.8-Max (xhigh)
& 93 & 91 & 70.2 & 1.71
& 104.90 & 72.98 & 1.59 & 906.20 & 0.46 \\

DeepSeek V4.1-Flash (max)
& 99 & 98 & 70.8 & 1.21
& 118.44 & 44.72 & 3.21 & 459.12 & 0.03 \\
\bottomrule
\end{tabular}%
}

\end{table*}

\begin{table}[t]
\centering
\begin{minipage}{.52\linewidth}
\centering
\arrayrulecolor{EvoRule}
\caption{Navigation performance beyond arrival confirmation.}
\label{tab:preconfirmation}
\begingroup
\footnotesize
\setlength{\tabcolsep}{2pt}
\renewcommand{\arraystretch}{1.15}

\resizebox{\linewidth}{!}{%
\begin{tabular}{@{}lccccc@{}}
\toprule
Benchmark
& OSR$\uparrow$ (\%)
& SR$\uparrow$ (\%)
& SPL$\uparrow$ (\%)
& NE$\downarrow$ (m)
& TL (m) \\
\midrule
R2R-CE
& 91.0 & 90.0 & 74.0 & 1.44 & 11.21 \\
RxR-CE
& 97.0 & 95.0 & 71.6 & 1.06 & 17.51 \\
HM3D-OVON
& 72.0 & 69.0 & 51.1 & 1.57 & 7.44 \\
HM3D v2
& 92.0 & 89.0 & 57.5 & 0.31 & 8.57 \\
\bottomrule
\end{tabular}%
}

\endgroup
\end{minipage}
\end{table}


\begin{figure}[!t]
\centering
\includegraphics[width=\linewidth]{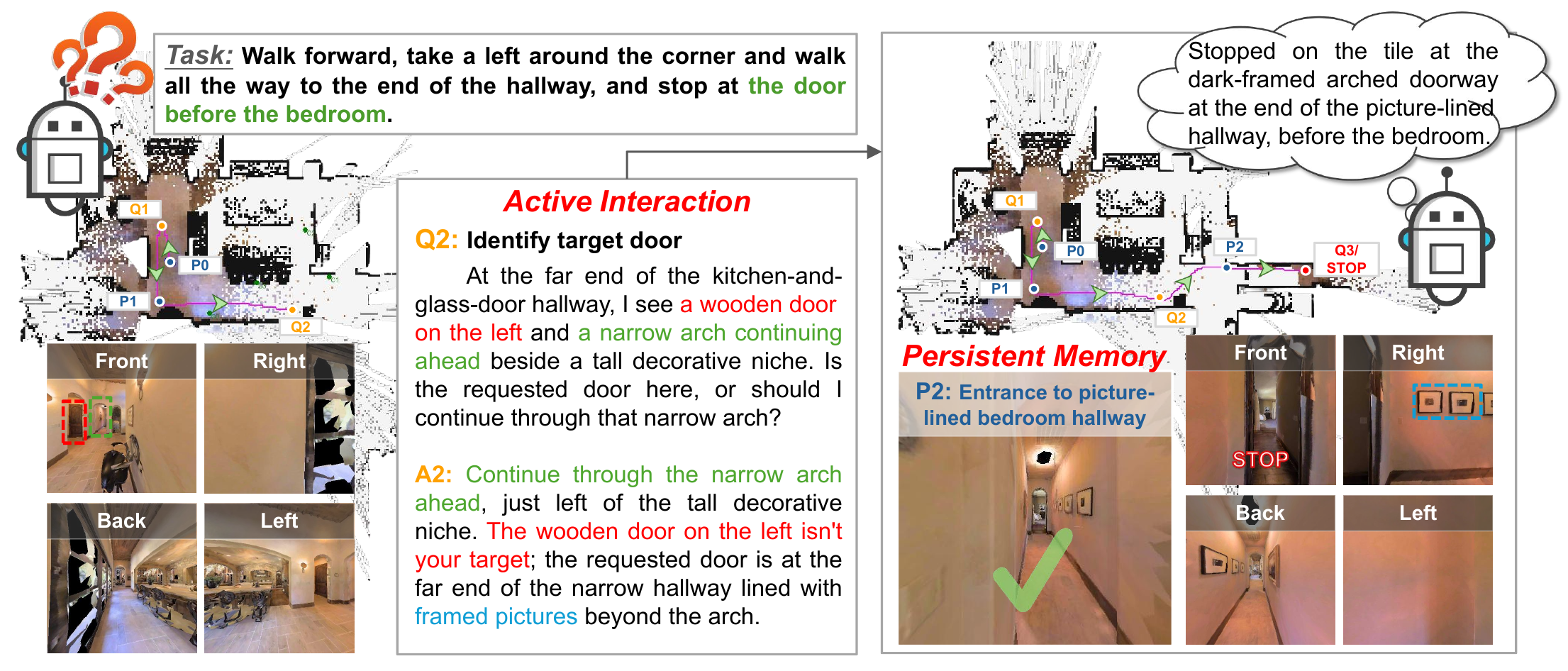}
\caption{Qualitative example of active interaction and persistent memory in R2R-CE. The agent actively asks whether a nearby wooden door is the intended destination or whether it should continue through a narrow archway. Based on the guide's response, the agent adjusts its navigation direction, proceeds through the archway, and records its entrance as place P2. Q2 and A2 denote the agent's second interaction question and the guide's corresponding answer, respectively, while P2 denotes the stored place record.}
\label{fig:case94}
\end{figure}

\subsection{Backbone Comparison}

\tabref{tab:backbone} compares the performance of GPT-6 Astra (ultra), Qwen3.8-Max (xhigh), and DeepSeek V4.1-Flash (max) on the assisted R2R-CE navigation task. The three models achieve success rates (SR) of 100\%, 91\%, and 98\%, respectively, with corresponding SPL scores of 82.2\%, 70.2\%, and 70.8\%. These results demonstrate that our unified tool-based framework can accommodate different decision backbones, with GPT-6 Astra achieving the best overall performance.

In terms of execution efficiency, the three models require an average of 15.57, 72.98, and 44.72 tool decisions, with navigator-active execution times of 164.08\,s, 906.20\,s, and 459.12\,s, respectively. Regarding tool selection behavior, GPT-6 Astra tends to favor waypoint navigation, enabling longer-distance movement with fewer decisions. In contrast, Qwen3.8-Max and DeepSeek V4.1-Flash adopt more conservative navigation strategies, relying more frequently on local discrete actions for movement and adjustment. Consequently, they require more tool decisions, resulting in substantially longer execution times.

For active interaction, the three models average 1.66, 1.59, and 3.21 questions per episode, respectively. DeepSeek V4.1-Flash requests assistance considerably more frequently than the other two models, indicating a greater reliance on external interaction for information and guidance during navigation. Considering only navigator-side API-equivalent costs, the average cost per episode is \$1.06, \$0.46, and \$0.03, respectively. Overall, GPT-6 Astra achieves the best performance in navigation success, path efficiency, and execution time, while DeepSeek V4.1-Flash maintains a high success rate at the lowest inference cost.

\subsection{Navigation Benefits Beyond Arrival Confirmation}

To investigate whether the benefits of active interaction extend beyond arrival confirmation, we conduct a further analysis that preserves other interactions but directly executes STOP upon the first arrival-confirmation-related request. This setting eliminates subsequent navigation adjustments guided by arrival-confirmation feedback, allowing us to examine the contributions of earlier interactions, such as directional guidance and target clarification.

As shown in \tabref{tab:preconfirmation}, our method still achieves strong navigation performance. Specifically, it obtains 90.0\% SR and 74.0\% SPL on R2R-CE, and 95.0\% SR and 71.6\% SPL on RxR-CE. Compared with the non-interactive R2R baseline (86.0\% SR and 63.7\% SPL), these results still demonstrate clear improvements.

On object navigation benchmarks, our method achieves 69.0\% SR and 51.1\% SPL on HM3D-OVON, and 89.0\% SR and 57.5\% SPL on HM3D v2. These results demonstrate that the guidance provided by active interaction during navigation is also highly effective. Meanwhile, the larger performance gaps on object navigation tasks highlight the value of further navigation adjustments under more stringent success criteria.


\begin{figure}[!t]
\centering
\includegraphics[width=\linewidth]{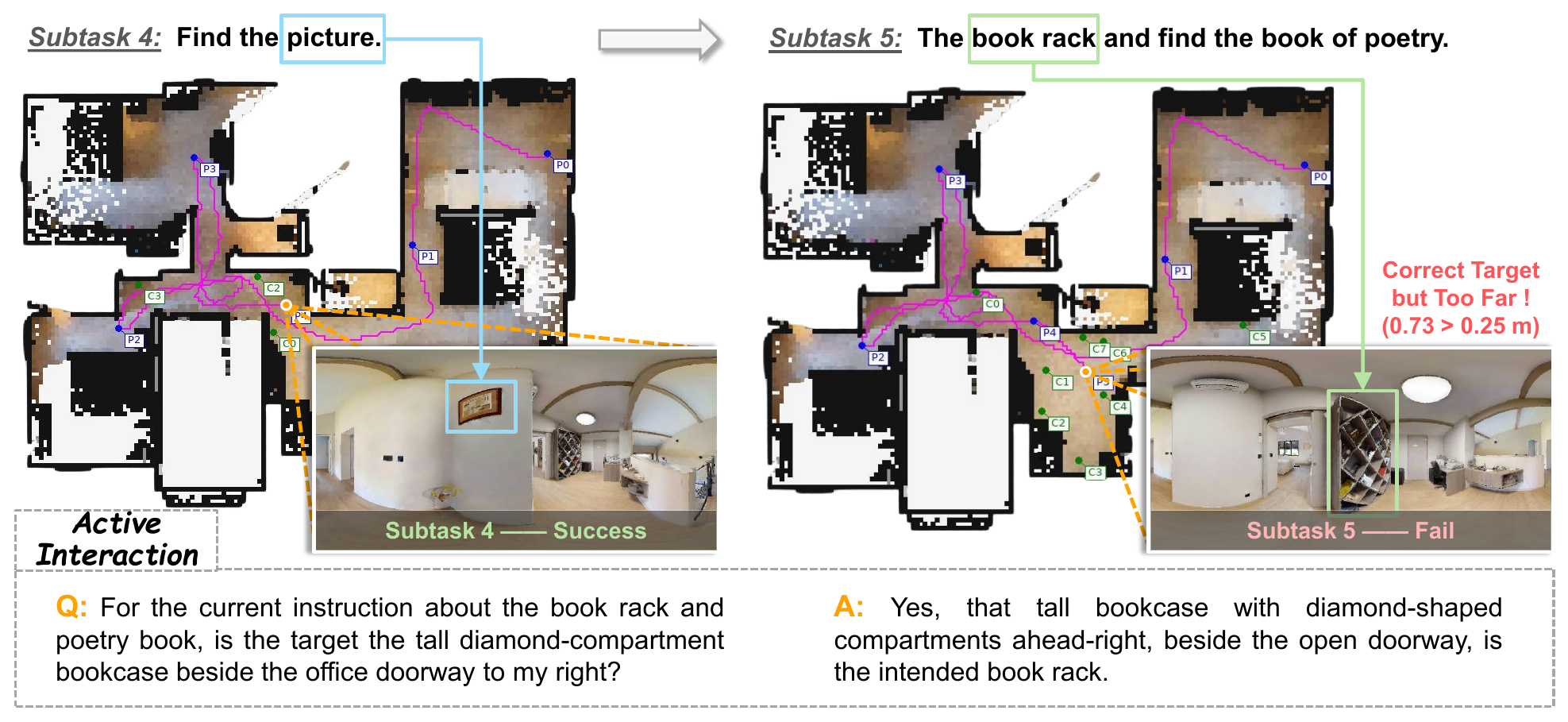}
\caption{Failure case analysis on GOAT-Bench. Left: the agent successfully completes the picture-finding subtask. Right: the agent subsequently identifies the bookcase with diamond-shaped compartments through active interaction, but stops 0.73\,m from the reference target, exceeding the required 0.25\,m success threshold. Orange dots on the maps indicate the agent's position. $P_i$ denotes a saved place in persistent memory. $C_i$ labels candidate waypoint $i$.}
\label{fig:failure}
\end{figure}

\subsection{Qualitative Visualization}

\figref{fig:case94} presents a visualization of a navigation example from R2R-CE. The task requires the agent to stop at the doorway before a bedroom. Upon reaching the end of a hallway connecting the kitchen and glass doors, the agent encounters a wooden door on the left and a narrow archway ahead. To resolve this ambiguity, the agent actively asks whether the wooden door is the intended destination or whether it should continue through the archway. The guide rules out the wooden door and indicates that the target lies at the end of a picture-lined hallway beyond the archway.

Based on this feedback, the agent adjusts its navigation direction, proceeds through the archway, and stores its entrance as a persistent place memory (P2), which can be retrieved to support future backtracking, route planning, and navigation across subsequent tasks.
This visualization illustrates how the agent resolves route ambiguity through active questioning, adjusts its navigation based on interaction feedback, and preserves key spatial information in persistent memory.

\subsection{Failure Case Analysis}
\label{app:failure}

We further examine all recorded failure cases in our experiments and find that the agent successfully identifies the correct target in every case, but fails to stop within the distance threshold specified by the benchmark. These results indicate that the observed failures primarily arise from inaccuracies in final positioning and stopping decisions rather than incorrect semantic target recognition.

\figref{fig:failure} illustrates a representative failure case from GOAT-Bench. In this task, the agent is instructed to locate a bookcase with diamond-shaped compartments. Through active interaction, the guide confirms the identity of the target bookcase, after which the agent localizes the target using RGB-D observations and executes navigation. Although the controller returns an arrival confirmation, the agent stops at a distance of 0.73\,m from the reference target location, exceeding the required success threshold of 0.25\,m. Consequently, the task is evaluated as a failure.

\begin{figure}[!t]
\centering
\includegraphics[width=\linewidth,height=.70\textheight,keepaspectratio]{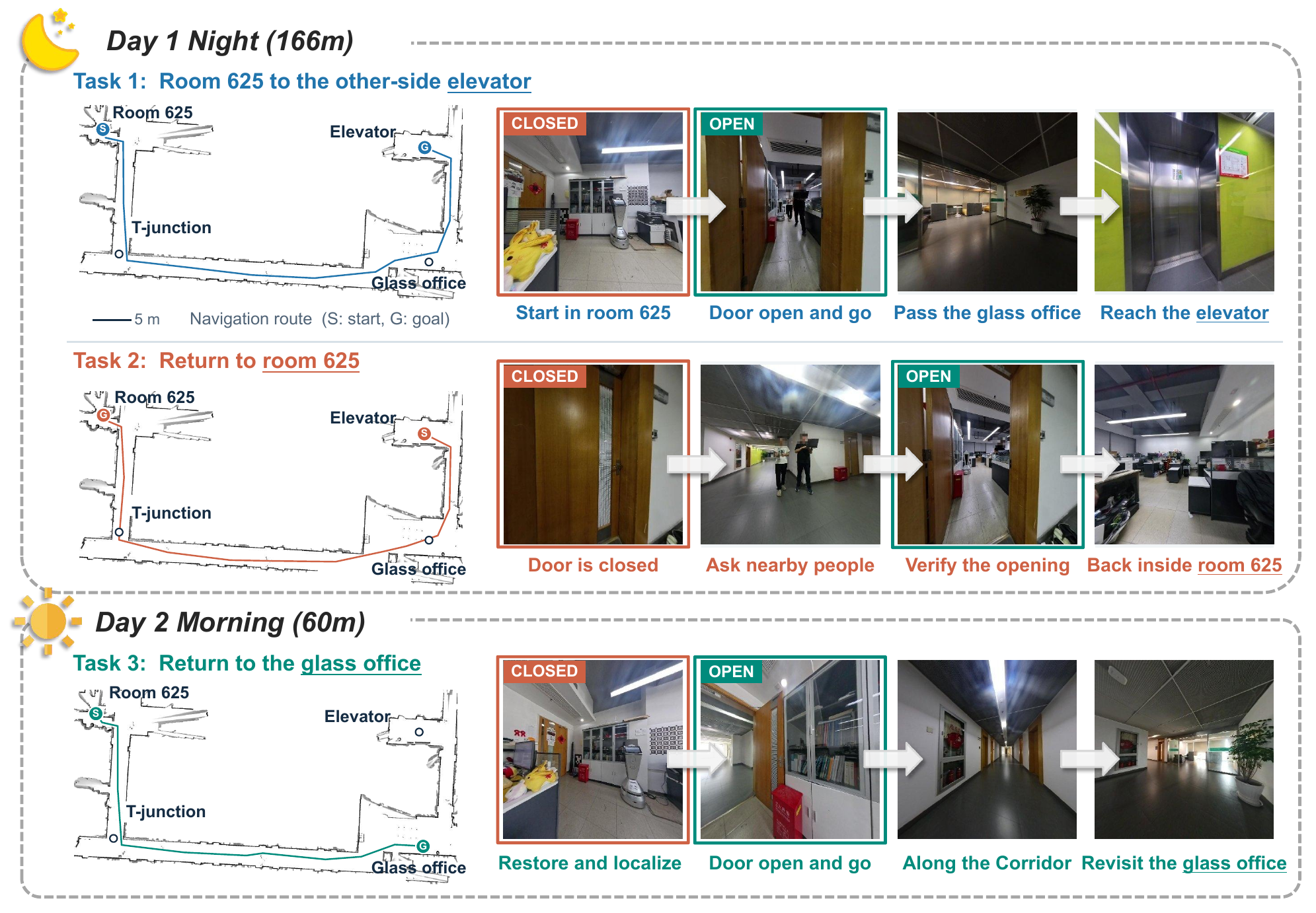}
\caption{Three consecutive real-world robot navigation tasks across two days. In Task 1, the robot navigates from room 625 to the elevator on the opposite side of the floor, recording key locations along the route. In Task 2, the robot returns to room 625 using persistent memory and actively requests human assistance when encountering a closed door. In Task 3, the robot identifies and restores the matching saved map on the following morning and navigates to the glass office visited the previous day. The three trajectories are visualized in a shared coordinate frame, demonstrating active interaction, adaptation to environmental changes, and cross-day memory reuse.}
\label{fig:real625}
\end{figure}

\subsection{Real-World Deployment}

\figref{fig:real625} illustrates three navigation tasks completed by the robot over two days in a real-world environment. The three navigation trajectories are derived from the robot's recorded poses in the map coordinate frame and overlaid on the final map saved on the first day.

On the first night, the robot receives a navigation instruction to
travel from room 625 to the elevator on the opposite side of the floor,
covering approximately $87\,\mathrm{m}$.
During navigation, the agent records several key locations, including a corridor junction, a fire-cabinet landmark, and a glass office beside a table-tennis table. Near the glass office, the agent actively asks for directional guidance, subsequently reaches the elevator, and saves the map state.

The user then instructs the robot to return to room 625. The agent retrieves and inspects the previously stored room record and maintains the interior of room 625 as its navigation destination. Notably, the door of room 625, which was open during departure, is closed upon return. To handle this environmental change, the agent actively requests assistance from a nearby person to open the door, verifies the changed passage through fresh visual observations, and resumes navigation to successfully complete the return task.

On the following morning, the user instructs the robot to navigate to
the glass office visited the previous day, covering approximately
$60\,\mathrm{m}$.
The agent first clarifies the intended destination through interaction, compares the current observations of room 625 with historical records, and selects the previously saved map. It then performs an in-place rotation to acquire observations from different orientations. After verifying its localization, the agent begins navigation. Upon reaching the glass office, despite differences in lighting conditions from the previous day, the agent successfully recognizes the location by matching current visual observations with historical records and completes the task based on actual arrival feedback.

These three successfully completed real-world navigation tasks demonstrate the capabilities of our method in long-term memory reuse and adaptation to dynamic environments. The agent can not only retain and reuse spatial memory across different tasks but also restore historical maps and recognize previously visited locations across days. By combining real-time observations with active interaction, it can further handle changes in door states and lighting conditions. These real-world experiments provide qualitative validation of the effectiveness of our unified framework integrating hybrid actions, active interaction, and persistent memory for lifelong embodied navigation.

\section{Discussion and Limitations}
\label{sec:discussion}

In this work, \method{} integrates hybrid actions, active interaction, and persistent memory into a unified embodied navigation framework, enabling agents to adaptively select movement strategies based on environmental conditions, actively seek external guidance, and reuse historical experience in subsequent tasks. Experimental results across eight navigation benchmarks demonstrate the applicability of our framework to vision-language navigation, object navigation, and multi-stage continuous navigation. Comparisons across different decision backbones further demonstrate the adaptability of the unified tool interface, while revealing distinct action selection strategies among models. In particular, GPT-6 Astra tends to favor waypoint navigation for longer-distance movement, whereas other models rely more frequently on local actions, highlighting the importance of appropriate action selection strategies for improving navigation efficiency.

Active interaction not only helps agents confirm arrival at a destination but also provides directional guidance, target clarification, and environmental assistance during navigation. Our analysis beyond arrival confirmation shows that the agent maintains strong navigation performance even without subsequent adjustments based on arrival-confirmation feedback, indicating that the benefits of active interaction extend beyond final success verification. Meanwhile, persistent memory enables the agent to associate interaction-derived information with spatial locations and reuse it in subsequent tasks. Memory-based revisit experiments demonstrate that the agent can maintain high success rates without additional interaction while reducing navigation steps and tool decisions. Real-world experiments further illustrate the agent's ability to restore historical maps, revisit known locations across days, and handle changes in door states through active interaction. These findings highlight the complementary roles of active interaction and long-term memory: interaction acquires new information, while memory allows that information to remain useful in subsequent tasks.

Despite the promising results, our method still faces limitations in real-time performance and operational cost. The navigation process involves multiple rounds of model inference, tool execution, and active interaction, resulting in decision latency and API costs that limit its applicability to long-term real-world robotic tasks. Future work will focus on optimizing decision efficiency and interaction strategies to reduce system latency and operational costs while maintaining navigation performance, advancing the framework toward more efficient and sustainable lifelong embodied navigation.

\section{Conclusion}
\label{sec:conclusion}

In this paper, we presented \method{}, a unified framework for lifelong embodied navigation that integrates hybrid actions, active interaction, and persistent memory. By enabling agents to adaptively select navigation actions, actively seek external guidance, and retain spatially grounded experience, our framework supports efficient navigation and knowledge reuse across tasks. Extensive experiments on eight navigation benchmarks demonstrate strong performance across vision-language navigation, object navigation, and multi-stage continuous navigation. Further analyses highlight the effectiveness of active interaction and the benefits of persistent memory in improving navigation efficiency during revisits. Real-world experiments additionally demonstrate the framework's ability to reuse historical experience across days and adapt to dynamic environmental changes. These results establish a promising foundation for embodied agents capable of continuously accumulating and reusing experience in lifelong navigation.

\FloatBarrier
\begingroup
\setlength{\bibsep}{3pt plus 1pt minus 1pt}
\bibliography{references/references}

@article{turing1950,
  author = {Turing, A. M.},
  title = {{Computing Machinery and Intelligence}},
  journal = {Mind},
  volume = {LIX},
  number = {236},
  pages = {433--460},
  year = {1950},
  doi = {10.1093/mind/LIX.236.433},
  url = {https://academic.oup.com/mind/article/LIX/236/433/986238},
}

@inproceedings{r2r2018,
  author = {Anderson, Peter and Wu, Qi and Teney, Damien and Bruce, Jake and Johnson, Mark and S{\"u}nderhauf, Niko and Reid, Ian and Gould, Stephen and van den Hengel, Anton},
  title = {{Vision-and-Language Navigation: Interpreting Visually-Grounded Navigation Instructions in Real Environments}},
  booktitle = {2018 IEEE/CVF Conference on Computer Vision and Pattern Recognition},
  pages = {3674--3683},
  year = {2018},
  doi = {10.1109/cvpr.2018.00387},
  url = {https://arxiv.org/abs/1711.07280},
  eprint = {1711.07280},
  archivePrefix = {arXiv},
}

@inproceedings{rxr2020,
  author = {Ku, Alexander and Anderson, Peter and Patel, Roma and Ie, Eugene and Baldridge, Jason},
  title = {{Room-Across-Room: Multilingual Vision-and-Language Navigation with Dense Spatiotemporal Grounding}},
  booktitle = {Proceedings of the 2020 Conference on Empirical Methods in Natural Language Processing (EMNLP)},
  pages = {4392--4412},
  year = {2020},
  doi = {10.18653/v1/2020.emnlp-main.356},
  url = {https://aclanthology.org/2020.emnlp-main.356/},
  eprint = {2010.07954},
  archivePrefix = {arXiv},
}

@inproceedings{vlnce2020,
  author = {Krantz, Jacob and Wijmans, Erik and Majumdar, Arjun and Batra, Dhruv and Lee, Stefan},
  title = {{Beyond the Nav-Graph: Vision-and-Language Navigation in Continuous Environments}},
  booktitle = {Computer Vision – ECCV 2020},
  pages = {104--120},
  year = {2020},
  doi = {10.1007/978-3-030-58604-1_7},
  url = {https://arxiv.org/abs/2004.02857},
  eprint = {2004.02857},
  archivePrefix = {arXiv},
}

@inproceedings{habitat2019,
  author = {Savva, Manolis and Kadian, Abhishek and Maksymets, Oleksandr and Zhao, Yili and Wijmans, Erik and Jain, Bhavana and Straub, Julian and Liu, Jia and Koltun, Vladlen and Malik, Jitendra and Parikh, Devi and Batra, Dhruv},
  title = {{Habitat: A Platform for Embodied AI Research}},
  booktitle = {2019 IEEE/CVF International Conference on Computer Vision (ICCV)},
  pages = {9338--9346},
  year = {2019},
  doi = {10.1109/iccv.2019.00943},
  url = {https://arxiv.org/abs/1904.01201},
  eprint = {1904.01201},
  archivePrefix = {arXiv},
}

@inproceedings{hm3d2021,
  author = {Ramakrishnan, Santhosh K. and Gokaslan, Aaron and Wijmans, Erik and Maksymets, Oleksandr and Clegg, Alex and Turner, John and Undersander, Eric and Galuba, Wojciech and Westbury, Andrew and Chang, Angel X. and Savva, Manolis and Zhao, Yili and Batra, Dhruv},
  title = {{Habitat-Matterport 3D Dataset (HM3D): 1000 Large-scale 3D Environments for Embodied AI}},
  booktitle = {Advances in Neural Information Processing Systems Datasets and Benchmarks Track},
  year = {2021},
  url = {https://aihabitat.org/datasets/hm3d/},
  eprint = {2109.08238},
  archivePrefix = {arXiv},
}

@inproceedings{hm3dsem2023,
  author = {Yadav, Karmesh and Ramrakhya, Ram and Ramakrishnan, Santhosh Kumar and Gervet, Theo and Turner, John and Gokaslan, Aaron and Maestre, Noah and Chang, Angel Xuan and Batra, Dhruv and Savva, Manolis and Clegg, Alexander William and Chaplot, Devendra Singh},
  title = {{Habitat-Matterport 3D Semantics Dataset}},
  booktitle = {2023 IEEE/CVF Conference on Computer Vision and Pattern Recognition (CVPR)},
  pages = {4927--4936},
  year = {2023},
  doi = {10.1109/cvpr52729.2023.00477},
  url = {https://aihabitat.org/datasets/hm3d-semantics/},
  eprint = {2210.05633},
  archivePrefix = {arXiv},
}

@inproceedings{ovon2024,
  author = {Yokoyama, Naoki and Ramrakhya, Ram and Das, Abhishek and Batra, Dhruv and Ha, Sehoon},
  title = {{HM3D-OVON: A Dataset and Benchmark for Open-Vocabulary Object Goal Navigation}},
  booktitle = {2024 IEEE/RSJ International Conference on Intelligent Robots and Systems (IROS)},
  pages = {5543--5550},
  year = {2024},
  doi = {10.1109/iros58592.2024.10802709},
  url = {https://arxiv.org/abs/2409.14296},
  eprint = {2409.14296},
  archivePrefix = {arXiv},
}

@inproceedings{goat2024,
  author = {Khanna, Mukul and Ramrakhya, Ram and Chhablani, Gunjan and Yenamandra, Sriram and Gervet, Theophile and Chang, Matthew and Kira, Zsolt and Chaplot, Devendra Singh and Batra, Dhruv and Mottaghi, Roozbeh},
  title = {{GOAT-Bench: A Benchmark for Multi-Modal Lifelong Navigation}},
  booktitle = {2024 IEEE/CVF Conference on Computer Vision and Pattern Recognition (CVPR)},
  pages = {16373--16383},
  year = {2024},
  doi = {10.1109/cvpr52733.2024.01549},
  url = {https://arxiv.org/abs/2404.06609},
  eprint = {2404.06609},
  archivePrefix = {arXiv},
}

@misc{vlln2025,
  author = {Huang, Wensi and Zhu, Shaohao and Wei, Meng and Zhang, Siqi and Xu, Jinming and Liu, Xihui and Wang, Hanqing and Wang, Tai and Zhao, Feng and Pang, Jiangmiao},
  title = {{VL-LN Bench: Towards Long-horizon Goal-oriented Navigation with Active Dialogs}},
  year = {2025},
  howpublished = {arXiv preprint arXiv:2512.22342},
  url = {https://arxiv.org/abs/2512.22342},
  eprint = {2512.22342},
  archivePrefix = {arXiv},
}

@inproceedings{coin2025,
  author = {Taioli, Francesco and Zorzi, Edoardo and Franchi, Gianni and Castellini, Alberto and Farinelli, Alessandro and Cristani, Marco and Wang, Yiming},
  title = {{Collaborative Instance Object Navigation: Leveraging Uncertainty-Awareness to Minimize Human-Agent Dialogues}},
  booktitle = {2025 IEEE/CVF International Conference on Computer Vision (ICCV)},
  pages = {18781--18792},
  year = {2025},
  doi = {10.1109/iccv51701.2025.01745},
  url = {https://arxiv.org/abs/2412.01250},
  eprint = {2412.01250},
  archivePrefix = {arXiv},
}

@inproceedings{lhpr2025,
  author = {Song, Xinshuai and Chen, Weixing and Liu, Yang and Chen, Weikai and Li, Guanbin and Lin, Liang},
  title = {{Towards Long-Horizon Vision-Language Navigation: Platform, Benchmark and Method}},
  booktitle = {2025 IEEE/CVF Conference on Computer Vision and Pattern Recognition (CVPR)},
  pages = {12078--12088},
  year = {2025},
  doi = {10.1109/cvpr52734.2025.01128},
  url = {https://openaccess.thecvf.com/content/CVPR2025/html/Song_Towards_Long-Horizon_Vision-Language_Navigation_Platform_Benchmark_and_Method_CVPR_2025_paper.html},
  eprint = {2412.09082},
  archivePrefix = {arXiv},
}

@inproceedings{fantasyvln2026,
  author = {Zuo, Jing and Mu, Lingzhou and Jiang, Fan and Ma, Chengcheng and Xu, Mu and Qi, Yonggang},
  title = {{FantasyVLN: Unified Multimodal Chain-of-Thought Reasoning for Vision-and-Language Navigation}},
  booktitle = {Proceedings of the IEEE/CVF Conference on Computer Vision and Pattern Recognition (CVPR)},
  pages = {29674--29683},
  year = {2026},
  url = {https://openaccess.thecvf.com/content/CVPR2026/html/Zuo_FantasyVLN_Unified_Multimodal_Chain-of-Thought_Reasoning_for_Vision-and-Language_Navigation_CVPR_2026_paper.html},
  eprint = {2601.13976},
  archivePrefix = {arXiv},
}

@inproceedings{ivln2023,
  author = {Krantz, Jacob and Banerjee, Shurjo and Zhu, Wang and Corso, Jason and Anderson, Peter and Lee, Stefan and Thomason, Jesse},
  title = {{Iterative Vision-and-Language Navigation}},
  booktitle = {2023 IEEE/CVF Conference on Computer Vision and Pattern Recognition (CVPR)},
  pages = {14921--14930},
  year = {2023},
  doi = {10.1109/cvpr52729.2023.01433},
  url = {https://arxiv.org/abs/2210.03087},
  eprint = {2210.03087},
  archivePrefix = {arXiv},
}

@article{etpnav2025,
  author = {An, Dong and Wang, Hanqing and Wang, Wenguan and Wang, Zun and Huang, Yan and He, Keji and Wang, Liang},
  title = {{ETPNav: Evolving Topological Planning for Vision-Language Navigation in Continuous Environments}},
  journal = {IEEE Transactions on Pattern Analysis and Machine Intelligence},
  volume = {47},
  number = {7},
  pages = {5130--5145},
  year = {2025},
  doi = {10.1109/tpami.2024.3386695},
  url = {https://arxiv.org/abs/2304.03047},
  eprint = {2304.03047},
  archivePrefix = {arXiv},
}

@article{memoir2026,
  author = {Xu, Yunzhe and Pan, Yiyuan and Liu, Zhe},
  title = {{Dream to Recall: Imagination-Guided Experience Retrieval for Memory-Persistent Vision-and-Language Navigation}},
  journal = {IEEE Transactions on Pattern Analysis and Machine Intelligence},
  volume = {48},
  number = {8},
  pages = {9035--9049},
  year = {2026},
  doi = {10.1109/tpami.2026.3679426},
  url = {https://arxiv.org/abs/2510.08553},
  eprint = {2510.08553},
  archivePrefix = {arXiv},
}

@article{navcot2025,
  author = {Lin, Bingqian and Nie, Yunshuang and Wei, Ziming and Chen, Jiaqi and Ma, Shikui and Han, Jianhua and Xu, Hang and Chang, Xiaojun and Liang, Xiaodan},
  title = {{NavCoT: Boosting LLM-Based Vision-and-Language Navigation via Learning Disentangled Reasoning}},
  journal = {IEEE Transactions on Pattern Analysis and Machine Intelligence},
  volume = {47},
  number = {7},
  pages = {5945--5957},
  year = {2025},
  doi = {10.1109/tpami.2025.3554559},
  url = {https://arxiv.org/abs/2403.07376},
  eprint = {2403.07376},
  archivePrefix = {arXiv},
}

@article{console2024,
  author = {Lin, Bingqian and Nie, Yunshuang and Wei, Ziming and Zhu, Yi and Xu, Hang and Ma, Shikui and Liu, Jianzhuang and Liang, Xiaodan},
  title = {{Correctable Landmark Discovery via Large Models for Vision-Language Navigation}},
  journal = {IEEE Transactions on Pattern Analysis and Machine Intelligence},
  volume = {46},
  number = {12},
  pages = {8534--8548},
  year = {2024},
  doi = {10.1109/tpami.2024.3407759},
  url = {https://arxiv.org/abs/2405.18721},
  eprint = {2405.18721},
  archivePrefix = {arXiv},
}

@inproceedings{navgpt2024,
  author = {Zhou, Gengze and Hong, Yicong and Wu, Qi},
  title = {{NavGPT: Explicit Reasoning in Vision-and-Language Navigation with Large Language Models}},
  booktitle = {Proceedings of the AAAI Conference on Artificial Intelligence},
  volume = {38},
  number = {7},
  pages = {7641--7649},
  year = {2024},
  doi = {10.1609/aaai.v38i7.28597},
  url = {https://arxiv.org/abs/2305.16986},
  eprint = {2305.16986},
  archivePrefix = {arXiv},
}

@misc{navgpt32026,
  author = {Zhou, Gengze and Hong, Yicong and Zhang, Jiazhao and Zhao, Xunyi and Zhou, Jian and Lei, Zixing and Wang, Zun and Zhao, Chongyang and Chen, Xionghui and Gould, Stephen and van den Hengel, Anton and Wu, Qi},
  title = {{NavGPT-3: Harnessing Context in a Hierarchical Navigation Runtime}},
  year = {2026},
  howpublished = {arXiv preprint arXiv:2610.10787},
  url = {https://arxiv.org/abs/2610.10787},
  eprint = {2610.10787},
  archivePrefix = {arXiv},
}

@misc{navharness2026,
  author = {Zhao, Xunyi and Zhou, Jian and Lin, Sihao and Zhou, Gengze and Li, Zerui and Yan, Xinyu and Liu, Jiajun and van den Hengel, Anton and Wu, Qi},
  title = {{NavHarness: Towards Lifelong Embodied Navigation}},
  year = {2026},
  howpublished = {arXiv preprint arXiv:2609.34276},
  url = {https://arxiv.org/abs/2609.34276},
  eprint = {2609.34276},
  archivePrefix = {arXiv},
}

@misc{harnessvln2026,
  author = {Chen, Yang and Che, Lirong and Huang, Zhenyu and Fu, Wenbo and Wang, Chuang and Cao, Xu and Liu, Daqi and Yang, Yuzhe and Su, Jian and Guo, Lan-Zhe},
  title = {{HarnessVLN: Unifying Training-Free Embodied Navigation through an Agent Harness}},
  year = {2026},
  howpublished = {arXiv preprint arXiv:2609.15195},
  url = {https://arxiv.org/abs/2609.15195},
  eprint = {2609.15195},
  archivePrefix = {arXiv},
}

@misc{supernav2026,
  author = {Zhang, Jinkai and Xu, Jingyi and Yu, Yuanhong and Guo, Jiarui and Hu, Ruizhen and Bao, Hujun and Zhou, Xiaowei and Peng, Sida},
  title = {{SuperNav: An Agentic Navigation System for Any Task in Any Scene}},
  year = {2026},
  howpublished = {arXiv preprint arXiv:2610.12126},
  url = {https://arxiv.org/abs/2610.12126},
  eprint = {2610.12126},
  archivePrefix = {arXiv},
}

@misc{asena2026,
  author = {Cheng, An-Chieh and Liu, Isabella and Bu, Edmund and Bjorck, Johan and Yin, Hongxu and Luo, Zhengyi and Kautz, Jan and Fan, Linxi "Jim" and Zhu, Yuke and Liu, Sifei},
  title = {{ASENA: Self-evolving Agents for Embodied Navigation}},
  year = {2026},
  howpublished = {arXiv preprint arXiv:2609.39207},
  url = {https://arxiv.org/abs/2609.39207},
  eprint = {2609.39207},
  archivePrefix = {arXiv},
}

@inproceedings{talk2escape2026,
  author = {Li, Zerui and Lin, Sihao and Shao, Yanyan and Zhang, Jiwen and Shi, Xiangyu and Li, Shijie and Wu, Qi},
  title = {{Talk2Escape: Conversational Grounding for Vision-and-Language Navigation}},
  booktitle = {2026 IEEE/RSJ International Conference on Intelligent Robots and Systems (IROS)},
  year = {2026},
  url = {https://arxiv.org/abs/2609.28296},
  eprint = {2609.28296},
  archivePrefix = {arXiv},
}

@misc{ggnav2026,
  author = {Zhu, Shaohao and Huang, Wensi and Zhang, Siqi and Yang, Yuqiang and Wei, Meng and Wan, Chenyang and Liu, Tianle and Wang, Tai and Xu, Jinming},
  title = {{GG-Nav: Goal-oriented Grounding Chain-of-Thought Elicits Reasoning in Navigation Foundation Models}},
  year = {2026},
  howpublished = {arXiv preprint arXiv:2610.14541},
  url = {https://arxiv.org/abs/2610.14541},
  eprint = {2610.14541},
  archivePrefix = {arXiv},
}

@inproceedings{procompnav2026,
  author = {Kwon, Junhyuk and Lee, Seungjoon and Park, Hyejin and Min, Kyle and Ok, Jungseul},
  title = {{ProCompNav: Proactive Instance Navigation with Comparative Judgment for Ambiguous User Queries}},
  booktitle = {Advances in Neural Information Processing Systems (NeurIPS)},
  year = {2026},
  note = {Accepted, oral presentation},
  url = {https://arxiv.org/abs/2605.06223},
  eprint = {2605.06223},
  archivePrefix = {arXiv},
}

@misc{roboatlas2026,
  author = {Schperberg, Alexander and Panda, Shivam K. and Vinod, Abraham P. and Rokaha, Bhagawan and Jawed, M. K. and Di Cairano, Stefano},
  title = {{RoboAtlas: Contextual Active SLAM}},
  year = {2026},
  howpublished = {arXiv preprint arXiv:2606.26046},
  url = {https://arxiv.org/abs/2606.26046},
  eprint = {2606.26046},
  archivePrefix = {arXiv},
}

@inproceedings{vlfm2024,
  author = {Yokoyama, Naoki and Ha, Sehoon and Batra, Dhruv and Wang, Jiuguang and Bucher, Bernadette},
  title = {{VLFM: Vision-Language Frontier Maps for Zero-Shot Semantic Navigation}},
  booktitle = {2024 IEEE International Conference on Robotics and Automation (ICRA)},
  pages = {42--48},
  year = {2024},
  doi = {10.1109/icra57147.2024.10610712},
  url = {https://arxiv.org/abs/2312.03275},
  eprint = {2312.03275},
  archivePrefix = {arXiv},
}

@inproceedings{conceptgraphs2024,
  author = {Gu, Qiao and Kuwajerwala, Ali and Morin, Sacha and Jatavallabhula, Krishna Murthy and Sen, Bipasha and Agarwal, Aditya and Rivera, Corban and Paul, William and Ellis, Kirsty and Chellappa, Rama and Gan, Chuang and de Melo, Celso Miguel and Tenenbaum, Joshua B. and Torralba, Antonio and Shkurti, Florian and Paull, Liam},
  title = {{ConceptGraphs: Open-Vocabulary 3D Scene Graphs for Perception and Planning}},
  booktitle = {2024 IEEE International Conference on Robotics and Automation (ICRA)},
  pages = {5021--5028},
  year = {2024},
  doi = {10.1109/icra57147.2024.10610243},
  url = {https://arxiv.org/abs/2309.16650},
  eprint = {2309.16650},
  archivePrefix = {arXiv},
}

@inproceedings{vlmaps2023,
  author = {Huang, Chenguang and Mees, Oier and Zeng, Andy and Burgard, Wolfram},
  title = {{Visual Language Maps for Robot Navigation}},
  booktitle = {2023 IEEE International Conference on Robotics and Automation (ICRA)},
  pages = {10608--10615},
  year = {2023},
  doi = {10.1109/icra48891.2023.10160969},
  url = {https://arxiv.org/abs/2210.05714},
  eprint = {2210.05714},
  archivePrefix = {arXiv},
}

@inproceedings{threedmem2025,
  author = {Yang, Yuncong and Yang, Han and Zhou, Jiachen and Chen, Peihao and Zhang, Hongxin and Du, Yilun and Gan, Chuang},
  title = {{3D-Mem: 3D Scene Memory for Embodied Exploration and Reasoning}},
  booktitle = {2025 IEEE/CVF Conference on Computer Vision and Pattern Recognition (CVPR)},
  pages = {17294--17303},
  year = {2025},
  doi = {10.1109/cvpr52734.2025.01612},
  url = {https://arxiv.org/abs/2411.17735},
  eprint = {2411.17735},
  archivePrefix = {arXiv},
}

@inproceedings{orion2024,
  author = {Dai, Yinpei and Peng, Run and Li, Sikai and Chai, Joyce},
  title = {{Think, Act, and Ask: Open-World Interactive Personalized Robot Navigation}},
  booktitle = {2024 IEEE International Conference on Robotics and Automation (ICRA)},
  pages = {3296--3303},
  year = {2024},
  doi = {10.1109/icra57147.2024.10610178},
  url = {https://arxiv.org/abs/2310.07968},
  eprint = {2310.07968},
  archivePrefix = {arXiv},
}

@inproceedings{cvdn2020,
  author = {Thomason, Jesse and Murray, Michael and Cakmak, Maya and Zettlemoyer, Luke},
  title = {{Vision-and-Dialog Navigation}},
  booktitle = {Proceedings of the Conference on Robot Learning},
  series = {Proceedings of Machine Learning Research},
  volume = {100},
  pages = {394--406},
  year = {2020},
  url = {https://proceedings.mlr.press/v100/thomason20a.html},
  eprint = {1907.04957},
  archivePrefix = {arXiv},
}

@inproceedings{hanna2019,
  author = {Nguyen, Khanh and Daum{\'e} III, Hal},
  title = {{Help, Anna! Visual Navigation with Natural Multimodal Assistance via Retrospective Curiosity-Encouraging Imitation Learning}},
  booktitle = {Proceedings of the 2019 Conference on Empirical Methods in Natural Language Processing and the 9th International Joint Conference on Natural Language Processing (EMNLP-IJCNLP)},
  pages = {684--695},
  year = {2019},
  doi = {10.18653/v1/d19-1063},
  url = {https://aclanthology.org/D19-1063/},
  eprint = {1909.01871},
  archivePrefix = {arXiv},
}

@inproceedings{ndtw2019,
  author       = {Gabriel Ilharco and Vihan Jain and Alexander Ku and Eugene Ie and Jason Baldridge},
  title        = {{General Evaluation for Instruction Conditioned Navigation using Dynamic Time Warping}},
  booktitle    = {Visually Grounded Interaction and Language (ViGIL), NeurIPS 2019 Workshop},
  year         = {2019},
  url          = {https://vigilworkshop.github.io/static/papers/33.pdf},
  bibsource    = {dblp computer science bibliography, https://dblp.org}
}

@misc{evaluation2018,
  author = {Anderson, Peter and Chang, Angel and Chaplot, Devendra Singh and Dosovitskiy, Alexey and Gupta, Saurabh and Koltun, Vladlen and Kosecka, Jana and Malik, Jitendra and Mottaghi, Roozbeh and Savva, Manolis and Zamir, Amir R.},
  title = {{On Evaluation of Embodied Navigation Agents}},
  year = {2018},
  howpublished = {arXiv preprint arXiv:1807.06757},
  url = {https://arxiv.org/abs/1807.06757},
  eprint = {1807.06757},
  archivePrefix = {arXiv},
}

@inproceedings{navid2024,
  author = {Zhang, Jiazhao and Wang, Kunyu and Xu, Rongtao and Zhou, Gengze and Hong, Yicong and Fang, Xiaomeng and Wu, Qi and Zhang, Zhizheng and Wang, He},
  title = {{NaVid: Video-based VLM Plans the Next Step for Vision-and-Language Navigation}},
  booktitle = {Robotics: Science and Systems XX},
  year = {2024},
  doi = {10.15607/rss.2024.xx.079},
  url = {https://arxiv.org/abs/2402.15852},
  eprint = {2402.15852},
  archivePrefix = {arXiv},
}

@inproceedings{navila2025,
  author = {Cheng, An-Chieh and Ji, Yandong and Yang, Zhaojing and Gongye, Zaitian and Zou, Xueyan and Kautz, Jan and B{\i}y{\i}k, Erdem and Yin, Hongxu and Liu, Sifei and Wang, Xiaolong},
  title = {{NaVILA: Legged Robot Vision-Language-Action Model for Navigation}},
  booktitle = {Robotics: Science and Systems XXI},
  year = {2025},
  doi = {10.15607/rss.2025.xxi.018},
  url = {https://arxiv.org/abs/2412.04453},
  eprint = {2412.04453},
  archivePrefix = {arXiv},
}

@inproceedings{streamvln2025,
  author = {Wei, Meng and Wan, Chenyang and Yu, Xiqian and Wang, Tai and Yang, Yuqiang and Mao, Xiaohan and Zhu, Chenming and Cai, Wenzhe and Wang, Hanqing and Chen, Yilun and Liu, Xihui and Pang, Jiangmiao},
  title = {{StreamVLN: Streaming Vision-and-Language Navigation via SlowFast Context Modeling}},
  booktitle = {2026 IEEE International Conference on Robotics and Automation (ICRA)},
  pages = {22174--22181},
  year = {2026},
  doi = {10.1109/icra57385.2026.11696190},
  url = {https://doi.org/10.1109/icra57385.2026.11696190},
  eprint = {2507.05240},
  archivePrefix = {arXiv},
}

@misc{qwenrobotnav2026,
  author = {Zhang, Jiazhao and Zhou, Gengze and Yin, Hale and Huang, Yiyang and Lei, Zixing and Peng, Qihang and Yuan, Haoqi and Zhang, Jie and Guo, Xudong and Chen, Xiaoyue and Yang, An and Huang, Fei and Yang, Zhibo and Lin, Junyang and Liu, Dayiheng and Zhou, Jingren and Yu, Zhuoyuan and Fan, Jingyang and Liang, Zhixuan and Lin, Pei and Wang, Ye and Li, Haoyang and Chen, Anzhe and Yan, Kun and Xu, Xiao and Li, Jiahao and Hu, Lulu and Zhang, Minying and Li, Shurui and Xiao, Wenhu and Bai, Shuai and Ren, Xuancheng and Lv, Chenxu and Wu, Chenfei and Chen, Xiong-Hui},
  title = {{Qwen-RobotNav Technical Report: A Scalable Navigation Model Designed for an Agentic Navigation System}},
  year = {2026},
  howpublished = {arXiv preprint arXiv:2606.18112},
  url = {https://arxiv.org/abs/2606.18112},
  eprint = {2606.18112},
  archivePrefix = {arXiv},
}

@misc{avertvln2026,
  author = {Liu, Minrui and Wang, Jingke and Huang, Yuehao and Su, Hao and Lv, Jiajun and Ma, Yukai and Liu, Yong},
  title = {{AVERT-VLN: Abstention-aware Visual Error Recovery and Training for Vision-and-Language Navigation}},
  year = {2026},
  howpublished = {arXiv preprint arXiv:2609.39579},
  url = {https://arxiv.org/abs/2609.39579},
  eprint = {2609.39579},
  archivePrefix = {arXiv},
}

@article{apexnav2025,
  author = {Zhang, Mingjie and Du, Yuheng and Wu, Chengkai and Zhou, Jinni and Qi, Zhenchao and Ma, Jun and Zhou, Boyu},
  title = {{ApexNAV: An Adaptive Exploration Strategy for Zero-Shot Object Navigation With Target-Centric Semantic Fusion}},
  journal = {IEEE Robotics and Automation Letters},
  volume = {10},
  number = {11},
  pages = {11530--11537},
  year = {2025},
  doi = {10.1109/lra.2025.3606388},
  url = {https://arxiv.org/abs/2504.14478},
  eprint = {2504.14478},
  archivePrefix = {arXiv},
}

@misc{aecnav2026,
  author = {Liu, Guanlin and Ling, Shaobin and Liu, Renyuan and Gong, Zeying and Hu, Junjie},
  title = {{AECNav: Active Evidence Consolidation for Efficient Zero-Shot Open-Vocabulary Object Navigation}},
  year = {2026},
  howpublished = {arXiv preprint arXiv:2608.10817},
  url = {https://arxiv.org/abs/2608.10817},
  eprint = {2608.10817},
  archivePrefix = {arXiv},
  primaryClass = {cs.RO},
}

@misc{p2dnav2026,
  author = {Sheng, Kai and Wang, Liuyi and Dai, Haojie and Li, Jinlong and Qin, Yongrui and He, Zongtao and Liu, Chengju and Chen, Qijun},
  title = {{P2DNav: Panorama-to-Downview Reasoning for Zero-shot Vision-and-Language Navigation}},
  year = {2026},
  howpublished = {arXiv preprint arXiv:2605.19634},
  url = {https://arxiv.org/abs/2605.19634},
  eprint = {2605.19634},
  archivePrefix = {arXiv},
}

@misc{navjev2026,
  author = {Sheng, Kai and Wang, Liuyi and Li, Jinlong and Dai, Haojie and Liu, Chengju and Chen, Qijun},
  title = {{NavJev: Efficient Vision-Language Navigation via Action-Centric Visual Compression and Discriminative Action-Semantic Memory}},
  year = {2026},
  howpublished = {arXiv preprint arXiv:2609.34969},
  url = {https://arxiv.org/abs/2609.34969},
  eprint = {2609.34969},
  archivePrefix = {arXiv},
}

@misc{clash2025,
  author = {Wang, Liuyi and He, Zongtao and Li, Jinlong and Xia, Ruihao and Hu, Mengxian and Yao, Chenpeng and Liu, Chengju and Tang, Yang and Chen, Qijun},
  title = {{CLASH: Collaborative Large-Small Hierarchical Framework for Continuous Vision-and-Language Navigation}},
  year = {2025},
  howpublished = {arXiv preprint arXiv:2512.10360},
  url = {https://arxiv.org/abs/2512.10360},
  eprint = {2512.10360},
  archivePrefix = {arXiv},
}

@article{wang2026comprehensive,
  author = {Wang, Liuyi and Sheng, Kai and He, Zongtao and Li, Jinlong and Qin, Yongrui and Dai, Haojie and Wang, Xiangyi and Yang, Jingwei and Yan, Qingqing and Liu, Chengju and Chen, Qijun},
  title = {{A Comprehensive Survey and Systematic Real-World Evaluation of Embodied Vision-and-Language Navigation}},
  journal = {IEEE Transactions on Automation Science and Engineering},
  volume = {23},
  pages = {14405--14434},
  year = {2026},
  publisher = {IEEE},
  doi = {10.1109/TASE.2026.3711284},
  url = {https://doi.org/10.1109/TASE.2026.3711284},
  eprint = {2607.09792},
  archivePrefix = {arXiv},
}

@article{mlanet2023,
  author = {He, Zongtao and Wang, Liuyi and Li, Shu and Yan, Qingqing and Liu, Chengju and Chen, Qijun},
  title = {{A Multilevel Attention Network with Sub-Instructions for Continuous Vision-and-Language Navigation}},
  journal = {Applied Intelligence},
  volume = {55},
  number = {7},
  pages = {657},
  year = {2025},
  doi = {10.1007/s10489-025-06544-9},
  url = {https://doi.org/10.1007/s10489-025-06544-9},
  eprint = {2303.01396},
  archivePrefix = {arXiv},
}

@inproceedings{uninavid2024,
  author = {Jiazhao Zhang and Kunyu Wang and Shaoan Wang and Minghan Li and Haoran Liu and Songlin Wei and Zhongyuan Wang and Zhizheng Zhang and He Wang},
  title = {{Uni-NaVid: A Video-based Vision-Language-Action Model for Unifying Embodied Navigation Tasks}},
  booktitle = {Robotics: Science and Systems XXI},
  year = {2025},
  doi = {10.15607/rss.2025.xxi.013},
  url = {https://doi.org/10.15607/rss.2025.xxi.013},
  eprint = {2412.06224},
  archivePrefix = {arXiv},
}

@inproceedings{janusvln2025,
  author = {Shuang Zeng and Dekang Qi and Xinyuan Chang and Feng Xiong and Shichao Xie and Xiaolong Wu and Shiyi Liang and Mu Xu and Xing Wei},
  title = {{JanusVLN: Decoupling Semantics and Spatiality with Dual Implicit Memory for Vision-Language Navigation}},
  booktitle = {International Conference on Learning Representations (ICLR)},
  year = {2026},
  url = {https://openreview.net/forum?id=RnuB0Nlbd5},
  eprint = {2509.22548},
  archivePrefix = {arXiv},
}

@InProceedings{decovln2026,
    author    = {Xin, Zihao and Li, Wentong and Jiang, Yixuan and Wang, Bin and Cong, Runmin and Qin, Jie and Huang, Shengjun},
    title     = {{DecoVLN: Decoupling Observation, Reasoning, and Correction for Vision-and-Language Navigation}},
    booktitle = {Proceedings of the IEEE/CVF Conference on Computer Vision and Pattern Recognition (CVPR)},
    month     = {June},
    year      = {2026},
    pages     = {32410-32420}
}

@inproceedings{activevln2025,
  author = {Zekai Zhang and Weiye Zhu and Hewei Pan and Xiangchen Wang and Rongtao Xu and Xing Sun and Feng Zheng},
  title = {{ActiveVLN: Towards Active Exploration via Multi-Turn RL in Vision-and-Language Navigation}},
  booktitle = {2026 IEEE International Conference on Robotics and Automation (ICRA)},
  pages = {12611--12618},
  year = {2026},
  doi = {10.1109/icra57385.2026.11696699},
  url = {https://doi.org/10.1109/icra57385.2026.11696699},
  eprint = {2509.12618},
  archivePrefix = {arXiv},
}

@misc{qwenvla2026,
  author = {Qiuyue Wang and Mingsheng Li and Jian Guan and Jinhui Ye and Sicheng Xie and Yitao Liu and Junhao Chen and Zhixuan Liang and Jie Zhang and Xintong Hu and Xuhong Huang and Pei Lin and Junyang Lin and Dayiheng Liu and Shuai Bai and Jingren Zhou and Jiazhao Zhang and Haoqi Yuan and Gengze Zhou and Hang Yin and Ye Wang and Yiyang Huang and Zixing Lei and Wujian Peng and Delin Chen and Yingming Zheng and Jingyang Fan and Xianwei Zhuang and Xin Zhou and Haoyang Li and Anzhe Chen and Tong Zhang and Xuejing Liu and Yuchong Sun and Ruizhe Chen and Zhaohai Li and Chenxu Lü and Zhibo Yang and Tao Yu and Xionghui Chen},
  title = {{Qwen-VLA: Unifying Vision-Language-Action Modeling across Tasks, Environments, and Robot Embodiments}},
  year = {2026},
  howpublished = {arXiv preprint arXiv:2605.30280},
  url = {https://arxiv.org/abs/2605.30280},
  eprint = {2605.30280},
  archivePrefix = {arXiv},
}

@misc{rynnbrain2026,
  author = {Ronghao Dang and Jiayan Guo and Bohan Hou and Sicong Leng and Kehan Li and Xin Li and Jiangpin Liu and Yunxuan Mao and Zhikai Wang and Yuqian Yuan and Minghao Zhu and Xiao Lin and Yang Bai and Qian Jiang and Yaxi Zhao and Minghua Zeng and Junlong Gao and Yuming Jiang and Jun Cen and Siteng Huang and Liuyi Wang and Wenqiao Zhang and Chengju Liu and Jianfei Yang and Shijian Lu and Deli Zhao},
  title = {{RynnBrain: Open Embodied Foundation Models}},
  year = {2026},
  howpublished = {arXiv preprint arXiv:2602.14979},
  url = {https://arxiv.org/abs/2602.14979},
  eprint = {2602.14979},
  archivePrefix = {arXiv},
}

@inproceedings{dualvln2025,
  author = {Meng Wei and Chenyang Wan and Jiaqi Peng and Xiqian Yu and Yuqiang Yang and Delin Feng and Wenzhe Cai and Chenming Zhu and Tai Wang and Jiangmiao Pang and Xihui Liu},
  title = {{Ground Slow, Move Fast: A Dual-System Foundation Model for Generalizable Vision-and-Language Navigation}},
  booktitle = {International Conference on Learning Representations (ICLR)},
  year = {2026},
  url = {https://openreview.net/pdf/1a25efebc28583e7d38570ab67050516fcc20d8e.pdf},
  eprint = {2512.08186},
  archivePrefix = {arXiv},
}

@misc{astranavworld2025,
  author = {Jintao Chen and Junjun Hu and Haochen Bai and Minghua Luo and Xinda Xue and Botao Ren and Chengyu Bai and Shichao Xie and Ziyi Chen and Fei Liu and Zedong Chu and Xiaolong Wu and Mu Xu and Shanghang Zhang},
  title = {{AstraNav-World: World Model for Foresight Control and Consistency}},
  year = {2025},
  howpublished = {arXiv preprint arXiv:2512.21714},
  url = {https://arxiv.org/abs/2512.21714},
  eprint = {2512.21714},
  archivePrefix = {arXiv},
}

@inproceedings{omninav2025,
  author = {Xinda Xue and Junjun Hu and Minghua Luo and Shichao Xie and Jintao Chen and Zixun Xie and Kuichen Quan and Wei Guo and Zedong Chu and Mu Xu and Zhengzhou Zhu},
  title = {{OmniNav: A Unified Framework for Prospective Exploration and Visual-Language Navigation}},
  booktitle = {International Conference on Learning Representations (ICLR)},
  year = {2026},
  url = {https://proceedings.iclr.cc/paper_files/paper/2026/hash/ad6363efb7af02f7db13d087c7e649bd-Abstract-Conference.html},
  eprint = {2509.25687},
  archivePrefix = {arXiv},
}

@misc{robostral2026,
  author = {Abdelaziz Bounhar and Abhijeet Somani and Aditi Kabra and Adrian Valente and Adrien Petralia and Adrien Sade and Alan Jeffares and Albert Jiang and Aleksandr Timashov and Alexandre Cahill and Alexandre Gavaudan and Alexandre Laval and Alexandre Sablayrolles and Amelie Heliou and Amos You and Andre Jonasson and Andrew Bai and Andrew Ehrenberg and Andrew Zhao and Angele Lenglemetz and Anmol Agarwal and Antonia Calvi and Arata Suzuki and Arjun Majumdar and Arthur Fournier and Artjom Joosen and Avinash Sooriyarachchi and Aylin Guliz Akkus and Aysenur Karaduman and Baptiste Bout and Baptiste Roziere and Baudouin De Monicault and Benjamin Holzschuh and Benjamin Lefaudeux and Benjamin Tibi and Bernhard Stadlbauer and Blazej Osinski and Camille Le Scao and Chaoran Yu and Charlotte Cronjager and Chen-Yo Sun and Chris Bamford and Christian Wallenwein and Christophe Renaudin and Clemence Lanfranchi and Corentin Barreau and Corentin Sautier and Cristiana-Diana Diaconu and Cyprien Courtot and Daniel Marczak and Darius Dabert and Diego de Las Casas and Dominik Nuss and Dylan Rubini and Dzmitry Soupel and Elizaveta Demyanenko and Elliot Chane-Sane and Emilien Fugier and Emmanuel Gottlob and Erik Aas and Etienne Goffinet and Etienne Millon and Eujeong Choi and Fabian Paischer and Fabian Schlager and Faruk Ahmed and Federico Baldassarre and Filip Szatkowski and Florian Wiesner and Gabrielle Berrada and Gaetan Ecrepont and Gaetan Lepage and Gaspard Blanchet and Gaspard Donada-Vidal and Gauthier Delerce and Gauthier Guinet and Genevieve Hayes and Georgii Novikov and Giada Pistilli and Gianluca Galletti and Guillaume Breton and Guillaume Kunsch and Guillaume Lample and Guillaume Martin and Guillaume Raille and Gunjan Dhanuka and Gunshi Gupta and Han Zhou and Harshil Shah and Hasan Furkan Vural and Hedi Hadiji and Hope McGovern and Hugo Cisneros and Hugo Thimonier and Indraneel Mukherjee and Ivan Cuevas Salazar and Jacques Sun and Jan Ludziejewski and Jason Rute and Jean Quentin and Jean-Hadrien Chabran and Jean-Malo Delignon and Jie Zhang and Joachim Studnia and Joep Barmentlo and Johannes Brandstetter and John Harvill and Jonas Amar and Jonas Schweizer and Josephine Delas and Josselin Somerville and Julien Denize and Julien Tauran and Kartik Khandelwal and Khyathi Raghavi Chandu and Kilian Tep and Kush Jain and Larissa Laich and Laura Calem and Laurence Aitchison and Laurent Callot and Laurent Fainsin and Leo Cotteleer and Leonard Blier and Lingxiao Zhao and Louis Martin and Louis Serrano and Lucile Saulnier and Ludovic Ho Fuh and Luis Montero and Maarten Buyl and Manon Chossegros and Marcin Mozejko and Margaret Jennings and Markus Hennerbichler and Martin Alexandre and Mathieu Poiree and Mathieu Schmitt and Mathilde Guillaumin and Matthieu Andre and Matthieu Dinot and Matthieu Futeral and Maurits Bleeker and Mauro Comi and Max Mynter and Maxim Berman and Maxime Darrin and Maxime Louis and Maximilian Augustin and Maximilian Muller and Melina Jingting Laimon and Mert Unsal and Mia Chiquier and Michael Pilcer and Michal Pietruszka and Michal Zajac and Mikhail Biriuchinskii and Minh-Quang Pham and Minwoo Kang and Morgane Riviere and Namit Katariya and Nathan Grinsztajn and Nathan Simpson and Neeraj Aggarwal and Neha Gupta and Ola Mysiak and Oliver Leicht and Olivier Bousquet and Olivier Duchenne and Parag Jain and Patricia Wang and Patrick Blies and Patrick von Platen and Paul Jacob and Paul Wambergue and Paula Kurylowicz and Pavan Kumar Reddy and Pavel Kuksa and Philippe Pinel and Philomene Chagniot and Pierre Stock and Pierre-Andre Savalle and Piotr Milos and Prateek Gupta and Pravesh Agrawal and Quentin Desreumaux and Quentin Torroba and Quercus Hernandez and Ram Ramrakhya and Randall Isenhour and Ranjit Parva and Raul Perez Pelaez and Reinhard Sonnleitner and Remi Delacourt and Richard Kurle and Rishi Shah and Rob Romijnders and Rohin Arora and Romain Sauvestre and Roman Soletskyi and Rosalie Millner and Rupert Menneer and Sagar Vaze and Samuel Barry and Samuel Belkadi and Samuel Humeau and Sanchit Gandhi and Sandeep Subramanian and Sarthak Mittal and Saskia Adaime and Sean Cha and Sebastian Kaltenbach and Shashwat Dalal and Shashwat Verma and Sherif Waly and Shrimai Prabhumoye and Siddhant Waghjale and Siddharth Gandhi and Simon Lepage and Simon Sorg and Soham Ghosh and Sophie Marbach and Srijan Mishra and Stanislas Lange and Steve Hong and Sumukh Aithal and Szymon Antoniak and Tarun Kumar Vangani and Teven Le Scao and Theo Cachet and Thibaut Lavril and Thomas Chabal and Thomas Coste and Thomas Defard and Thomas Foubert and Thomas Robert and Thomas Wang and Tianyu Zhang and Tim Lawson and Timothee Lacroix and Tobias Kronlachner and Tom Bewley and Tom Edwards and Tomas Hodan and Tuhin Das and Tyler Wang and Ulrick BLE and Umar Jamil and Umberto Tomasini and Valentin Mace and Van Phung and Vedant Nanda and Victor Jouault and Victor Letzelter and Victor Paltz and Victor Poucheret and Vincent Maladiere and Vincent Pfister and Virgile Richard and Vladislav Bataev and Wassim Bouaziz and Wen Ding Li and William Havard and William Marshall and Xinghui Li and Xingran Guo and Xinyu Yang and Yann Dreze and Yassine El Ouahidi and Yassir Bendou and Yihan Wang and Yimu Pan and Yves Martin des Taillades and Zaccharie Ramzi and Zhenlin Xu and Zsofia Csakany},
  title = {{Robostral Navigate}},
  year = {2026},
  howpublished = {arXiv preprint arXiv:2607.20785},
  url = {https://arxiv.org/abs/2607.20785},
  eprint = {2607.20785},
  archivePrefix = {arXiv},
}

@inproceedings{sgnav2024,
 author = {Yin, Hang and Xu, Xiuwei and Wu, Zhenyu and Zhou, Jie and Lu, Jiwen},
 booktitle = {Advances in Neural Information Processing Systems},
 doi = {10.52202/079017-0171},
 editor = {A. Globerson and L. Mackey and D. Belgrave and A. Fan and U. Paquet and J. Tomczak and C. Zhang},
 pages = {5285--5307},
 publisher = {Curran Associates, Inc.},
 title = {{SG-Nav: Online 3D Scene Graph Prompting for LLM-based Zero-shot Object Navigation}},
 url = {https://proceedings.neurips.cc/paper_files/paper/2024/file/098491b37deebbe6c007e69815729e09-Paper-Conference.pdf},
 volume = {37},
 year = {2024}
}

@inproceedings{mtu3d2025,
  author = {Ziyu Zhu and Xilin Wang and Yixuan Li and Zhuofan Zhang and Xiaojian Ma and Yixin Chen and Baoxiong Jia and Wei Liang and Qian Yu and Zhidong Deng and Siyuan Huang and Qing Li},
  title = {{Move to Understand a 3D Scene: Bridging Visual Grounding and Exploration for Efficient and Versatile Embodied Navigation}},
  booktitle = {Proceedings of the IEEE/CVF International Conference on Computer Vision (ICCV)},
  pages = {8120--8132},
  year = {2025},
  url = {https://openaccess.thecvf.com/content/ICCV2025/html/Zhu_Move_to_Understand_a_3D_Scene_Bridging_Visual_Grounding_and_ICCV_2025_paper.html},
  eprint = {2507.04047},
  archivePrefix = {arXiv},
}

@inproceedings{navfom2025,
 author = {Zhang, Jiazhao and Li, Anqi and Qi, Yunpeng and Li, Minghan and Liu, Jiahang and Wang, Shaoan and Liu, Haoran and Zhou, Gengze and Wu, Yuze and Li, Xingxing and Fan, Yuxin and Li, Wenjun and Chen, Zhibo and Gao, Fei and Wu, Qi and Zhang, Zhizheng and Wang, He},
 booktitle = {International Conference on Learning Representations (ICLR)},
 editor = {C. Vondrick and B. Hariharan and C. Raffel and L. Pinto and D. Yang and A. Faust},
 pages = {127293--127322},
 title = {{Embodied Navigation Foundation Model}},
 url = {https://proceedings.iclr.cc/paper_files/paper/2026/file/ceb45c48e29e7f49b0b47edb98e43691-Paper-Conference.pdf},
 volume = {2026},
 year = {2026}
}

@misc{abotn02026,
  author = {Zedong Chu and Shichao Xie and Xiaolong Wu and Yanfen Shen and Minghua Luo and Zhengbo Wang and Fei Liu and Xiaoxu Leng and Junjun Hu and Mingyang Yin and Jia Lu and Yingnan Guo and Kai Yang and Jiawei Han and Xu Chen and Yanqing Zhu and Yuxiang Zhao and Xin Liu and Yirong Yang and Ye He and Jiahang Wang and Yang Cai and Tianlin Zhang and Li Gao and Liu Liu and Mingchao Sun and Fan Jiang and Chiyu Wang and Zhicheng Liu and Hongyu Pan and Honglin Han and Zhining Gu and Kuan Yang and Jianfang Zhang and Di Jing and Zihao Guan and Wei Guo and Guoqing Liu and Di Yang and Xiangpo Yang and Menglin Yang and Hongguang Xing and Weiguo Li and Mu Xu},
  title = {{ABot-N0: Technical Report on the VLA Foundation Model for Versatile Embodied Navigation}},
  year = {2026},
  howpublished = {arXiv preprint arXiv:2602.11598},
  url = {https://arxiv.org/abs/2602.11598},
  eprint = {2602.11598},
  archivePrefix = {arXiv},
}

@inproceedings{memoryexplorer2026,
  author = {Sen Wang and Bangwei Liu and Zhenkun Gao and Lizhuang Ma and Xuhong Wang and Yuan Xie and Xin Tan},
  title = {{Explore with Long-term Memory: A Benchmark and Multimodal LLM-based Reinforcement Learning Framework for Embodied Exploration}},
  booktitle = {Proceedings of the IEEE/CVF Conference on Computer Vision and Pattern Recognition (CVPR)},
  pages = {37098--37108},
  year = {2026},
  url = {https://openaccess.thecvf.com/content/CVPR2026/html/Wang_Explore_with_Long-term_Memory_A_Benchmark_and_Multimodal_LLM-based_Reinforcement_CVPR_2026_paper.html},
  eprint = {2601.10744},
  archivePrefix = {arXiv},
}

@misc{evomemnav2026,
  author = {Zuhao Ge and Xiaosong Jia and Chao Wu and Yuchen Zhou and Zuxuan Wu and Yu-Gang Jiang},
  title = {{EvoMemNav: Efficient Self-Evolving Fine-Grained Memory for Zero-Shot Embodied Navigation}},
  year = {2026},
  howpublished = {arXiv preprint arXiv:2606.03509},
  url = {https://arxiv.org/abs/2606.03509},
  eprint = {2606.03509},
  archivePrefix = {arXiv},
}

@inproceedings{astranavmemory2025,
  author = {Junjun Hu and Xinda Xue and Botao Ren and Minghua Luo and Jintao Chen and Haochen Bai and Liangliang You and Mu Xu},
  title = {{AstraNav-Memory: Contexts Compression for Long Memory}},
  booktitle = {Proceedings of the IEEE/CVF Conference on Computer Vision and Pattern Recognition (CVPR)},
  pages = {8097--8109},
  year = {2026},
  url = {https://openaccess.thecvf.com/content/CVPR2026/html/Hu_AstraNav-Memory_Contexts_Compression_for_Long_Memory_CVPR_2026_paper.html},
  eprint = {2512.21627},
  archivePrefix = {arXiv},
}

@misc{stegnav2026,
  author = {Yang Chen and Zhenyu Huang and Wenbo Fu and Danyang Peng and Shi-Yu Tian and Kun-Yang Yu and Lan-Zhe Guo},
  title = {{STEGNav: Spatio-Temporal Event Graph Reasoning for Multimodal Lifelong Object Navigation}},
  year = {2026},
  howpublished = {arXiv preprint arXiv:2608.28279},
  url = {https://arxiv.org/abs/2608.28279},
  eprint = {2608.28279},
  archivePrefix = {arXiv},
}

@misc{gsmem2026,
  author = {Yiren Lu and Yi Du and Disheng Liu and Yunlai Zhou and Chen Wang and Yu Yin},
  title = {{GSMem: 3D Gaussian Splatting as Persistent Spatial Memory for Zero-Shot Embodied Exploration and Reasoning}},
  year = {2026},
  howpublished = {arXiv preprint arXiv:2603.19137},
  url = {https://arxiv.org/abs/2603.19137},
  eprint = {2603.19137},
  archivePrefix = {arXiv},
}

@misc{metanav2026,
  author = {Xueying Li and Feng Lyu and Hao Wu and Mingliu Liu and Jia-Nan Liu and Guozi Liu},
  title = {{Stop Wandering: Efficient Vision-Language Navigation via Metacognitive Reasoning}},
  year = {2026},
  howpublished = {arXiv preprint arXiv:2604.02318},
  url = {https://arxiv.org/abs/2604.02318},
  eprint = {2604.02318},
  archivePrefix = {arXiv},
}

@misc{obsgraph2026,
  author = {Taekbeom Lee and Youngseok Jang and Jeonghwa Heo and Jeongjun Choi and H. Jin Kim},
  title = {{ObsGraph: Hierarchical Observation Representation for Embodied Reasoning and Exploration}},
  year = {2026},
  howpublished = {arXiv preprint arXiv:2606.24068},
  url = {https://arxiv.org/abs/2606.24068},
  eprint = {2606.24068},
  archivePrefix = {arXiv},
}

@inproceedings{navillm2023,
  author = {Duo Zheng and Shijia Huang and Lin Zhao and Yiwu Zhong and Liwei Wang},
  title = {{Towards Learning a Generalist Model for Embodied Navigation}},
  booktitle = {Proceedings of the IEEE/CVF Conference on Computer Vision and Pattern Recognition (CVPR)},
  pages = {13624--13634},
  year = {2024},
  url = {https://openaccess.thecvf.com/content/CVPR2024/html/Zheng_Towards_Learning_a_Generalist_Model_for_Embodied_Navigation_CVPR_2024_paper.html},
  eprint = {2312.02010},
  archivePrefix = {arXiv},
}

@inproceedings{auxthink2025,
 author = {Wang, Shuo and Wang, Yongcai and Li, Wanting and Cai, Xudong and Wang, Yucheng and Chen, Maiyue and Wang, Kaihui and Su, Zhizhong and Li, Deying and Fan, Zhaoxin},
 booktitle = {Advances in Neural Information Processing Systems},
 doi = {10.52202/085713-1040},
 editor = {D. Belgrave and C. Zhang and H. Lin and R. Pascanu and P. Koniusz and M. Ghassemi and N. Chen},
 pages = {30892--30916},
 publisher = {Curran Associates, Inc.},
 title = {{Aux-Think: Exploring Reasoning Strategies for Data-Efficient Vision-Language Navigation}},
 url = {https://proceedings.neurips.cc/paper_files/paper/2025/file/2c90bf5bffe6497730ecade3a3a37458-Paper-Conference.pdf},
 volume = {38, Main Conference},
 year = {2025}
}

@misc{internvlan12025,
  author = {{Intern Robotics}},
  title = {{InternVLA-N1: An Open Dual-System Vision-Language Navigation Foundation Model with Learned Latent Plans}},
  year = {2025},
  note = {Technical report, Shanghai AI Laboratory},
  url = {https://internrobotics.github.io/internvla-n1.github.io/static/pdfs/InternVLA_N1.pdf},
}

@inproceedings{hgr2026,
author={Chen, Peixin and Zhang, Guoxi and Ma, Jianwei and Li, Qing},
title={{Hypothesis Graph Refinement: Hypothesis-Driven Exploration with Cascade Error Correction for Embodied Navigation}},
booktitle={Computer Vision -- ECCV 2026},
year={2026},
publisher={Springer Nature Switzerland},
address={Cham},
pages={556--573},
}

@misc{wang2026lightnav,
      title={{LightNav-0: Eliciting VLM Spatial Intelligence for Generalist Embodied Navigation}}, 
      author={Shaoan Wang and Aocheng Luo and Fei Huang and Jingyi Xu and Xiaoyang Wang and Yueyu Wang and Qianli Ma and Fan Yang and Ran Mei and Jia Wei and Jiangpeng Hu and Xuhao Liu and Hongming Chen and Yuanbin Shao and Yiyang Lin and Ziliang Li and Liang Pan and Xinhang Liu and Yuntao Ma and Tingxiang Fan},
      year={2026},
      eprint={2608.30935},
      archivePrefix={arXiv},
      primaryClass={cs.RO},
      url={https://arxiv.org/abs/2608.30935}, 
}

@misc{himm2026,
      title={{HIMM: Human-Inspired Long-Term Memory Modeling for Embodied Exploration and Question Answering}}, 
      author={Ji Li and Bo Wang and Jing Xia and Mingyi Li and Shiyan Hu},
      year={2026},
      eprint={2602.15513},
      archivePrefix={arXiv},
      primaryClass={cs.RO},
      url={https://arxiv.org/abs/2602.15513}, 
}

@inproceedings{ssmgnav2026,
  author={Niu, Haochen and Zhang, Lantao and Ji, Xingwu and Ying, Rendong and Liu, Peilin and Wen, Fei},
  booktitle={2026 IEEE International Conference on Robotics and Automation (ICRA)}, 
  title={{SSMG-Nav: Enhancing Lifelong Object Navigation with Semantic Skeleton Memory Graph}}, 
  year={2026},
  volume={},
  number={},
  pages={2258-2265},
  doi={10.1109/ICRA57385.2026.11697433}}

@article{reexplore2025,
  title={{ReEXplore: Improving MLLMs for Embodied Exploration with Contextualized Retrospective Experience Replay}},
  author={Zhang, Gengyuan and Ding, Mingcong and Wu, Jingpei and Liao, Ruotong and Tresp, Volker},
  journal={arXiv preprint arXiv:2511.19033},
  year={2025}
}

@article{dai2026dped,
  title={DPed-VLN: A Benchmark for Socially Compliant Vision-and-Language Navigation in Dynamic Pedestrian Environments},
  author={Dai, Haojie and Wang, Xiangyi and Wang, Liuyi and Sheng, Kai and He, Zongtao and Liu, Chengju and Ye, Wei and Chen, Qijun},
  journal={arXiv preprint arXiv:2609.21504},
  year={2026}
}

@INPROCEEDINGS{wang2024causal,
  author={Wang, Liuyi and He, Zongtao and Dang, Ronghao and Shen, Mengjiao and Liu, Chengju and Chen, Qijun},
  booktitle={2024 IEEE/CVF Conference on Computer Vision and Pattern Recognition (CVPR)}, 
  title={Vision-and-Language Navigation via Causal Learning}, 
  year={2024},
  volume={},
  number={},
  pages={13139-13150},
  doi={10.1109/CVPR52733.2024.01248}}

@article{wang2026magic,
  title={Magic: Meta-ability guided interactive chain-of-distillation for effective-and-efficient vision-and-language navigation},
  author={Wang, Liuyi and He, Zongtao and Shen, Mengjiao and Yang, Jingwei and Liu, Chengju and Chen, Qijun},
  journal={IEEE Transactions on Pattern Analysis and Machine Intelligence},
  year={2026},
  publisher={IEEE}
}

@inproceedings{wang2025rethinking,
  title={Rethinking the embodied gap in vision-and-language navigation: A holistic study of physical and visual disparities},
  author={Wang, Liuyi and Xia, Xinyuan and Zhao, Hui and Wang, Hanqing and Wang, Tai and Chen, Yilun and Liu, Chengju and Chen, Qijun and Pang, Jiangmiao},
  booktitle={2025 IEEE/CVF International Conference on Computer Vision (ICCV)},
  pages={9455--9465},
  year={2025},
  organization={IEEE}
}

@inproceedings{yu2026correctnav,
  title={Correctnav: Self-correction flywheel empowers vision-language-action navigation model},
  author={Yu, Zhuoyuan and Long, Yuxing and Yang, Zihan and Zeng, Chengyan and Fan, Hongwei and Zhang, Jiyao and Dong, Hao},
  booktitle={Proceedings of the AAAI Conference on Artificial Intelligence},
  volume={40},
  number={22},
  pages={18737--18745},
  year={2026}
}

@misc{openai2026gpt6astra,
  title  = {{GPT-6 Astra System Card}},
  author = {{OpenAI}},
  year   = {2026},
  month  = sep,
  url    = {https://deploymentsafety.openai.com/gpt-6-astra}
}

@misc{qwen2026qwen38max,
  title  = {{Qwen3.8-Max}: A New Bar for Coding and Cowork},
  author = {{Qwen Team}},
  year   = {2026},
  month  = aug,
  url    = {https://qwen.ai/blog?id=qwen3.8}
}

@misc{deepseek2026v41flash,
  title  = {{DeepSeek-V4.1-Flash}: Pushing the Limits of {KV} Cache Compression},
  author = {{DeepSeek-AI}},
  year   = {2026},
  url    = {https://huggingface.co/deepseek-ai/DeepSeek-V4.1-Flash/blob/main/DeepSeek_V41_Tech_Report.pdf}
}
\endgroup
\clearpage
\begingroup
\small
\setlength{\parskip}{4pt}
\appendix
\setcounter{figure}{0}
\renewcommand{\thefigure}{S\arabic{figure}}

\clearpage
\section{Additional Qualitative Examples}
\label{app:more-qualitative}

We present one example from each of the eight benchmarks to examine how
\method{} combines navigation, interaction, and memory in different task
settings. Each figure pairs selected observed maps and panoramic views with
the corresponding tool requests. The displayed decisions are excerpts from
the recorded executions; intermediate movements and tool calls are omitted.
The single-task examples illustrate route following, object approach, and
instance disambiguation. The continuous-task examples retain the spatial
context while the current navigation goal changes.

\subsection{R2R-CE: Traversing Successive Rooms}
\label{app:qual-r2r}

\begin{figure}[!htbp]
\centering
\includegraphics[width=\linewidth]{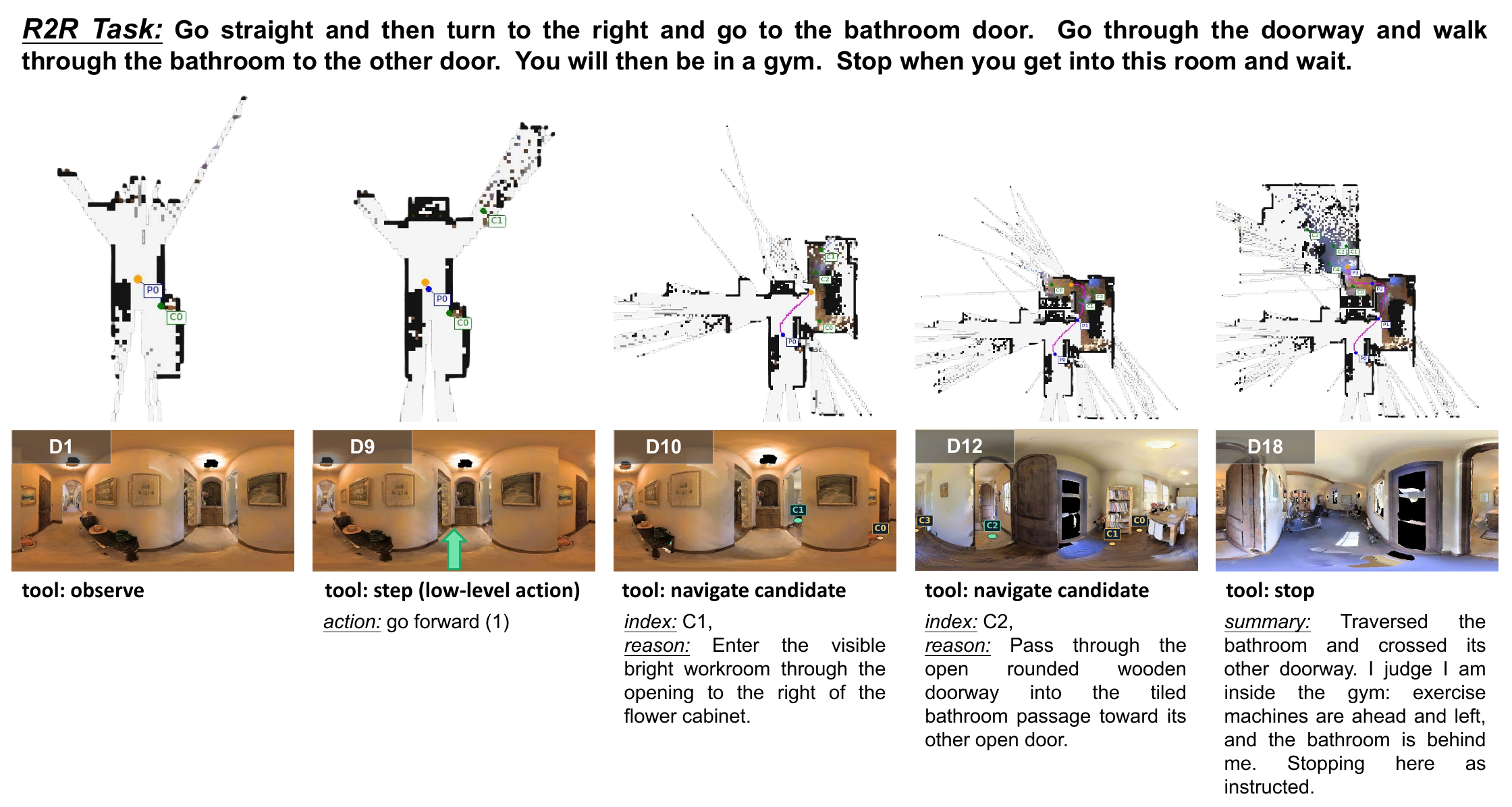}
\caption{\textbf{Hybrid navigation on R2R-CE.}
The instruction requires passing through a bathroom to reach a gym.
The observed map expands as the agent moves from the initial hallway through
a bright workroom and the tiled bathroom. \tool{step} and
\tool{navigate\_candidate} requests provide complementary ways to expose and
traverse the connecting doorways. The final view contains exercise equipment,
and an explicit \tool{stop} submits the completed task. Orange dots on the maps indicate the agent's position after executing the
corresponding decision. $P_i$ denotes a saved place in persistent memory. In the panoramas, $C_i$ labels candidate waypoint $i$,
and green markers indicate the selected waypoint or pixel. $D_i$ denotes the $i$-th tool-level decision.}
\label{fig:qual-r2r}
\end{figure}
\FloatBarrier

\figref{fig:qual-r2r} illustrates how the agent grounds an instruction whose
destination is reached through several intermediate rooms. Local turning and
forward movement reveal the opening beside the flower cabinet, after which
a candidate request enters the bright workroom. A later view exposes the
rounded wooden doorway into the bathroom, allowing the agent to continue
through the tiled passage toward the exercise room. The accumulated map
connects these successive observations, while the saved place descriptions
associate the route with recognizable room entrances. At the endpoint, the
agent's stopping summary identifies the gym through the exercise machines
ahead and the bathroom behind it.

\clearpage
\subsection{RxR-CE: Grounding a Detailed Route in Landmarks}
\label{app:qual-rxr}

\begin{figure}[!htbp]
\centering
\includegraphics[width=\linewidth]{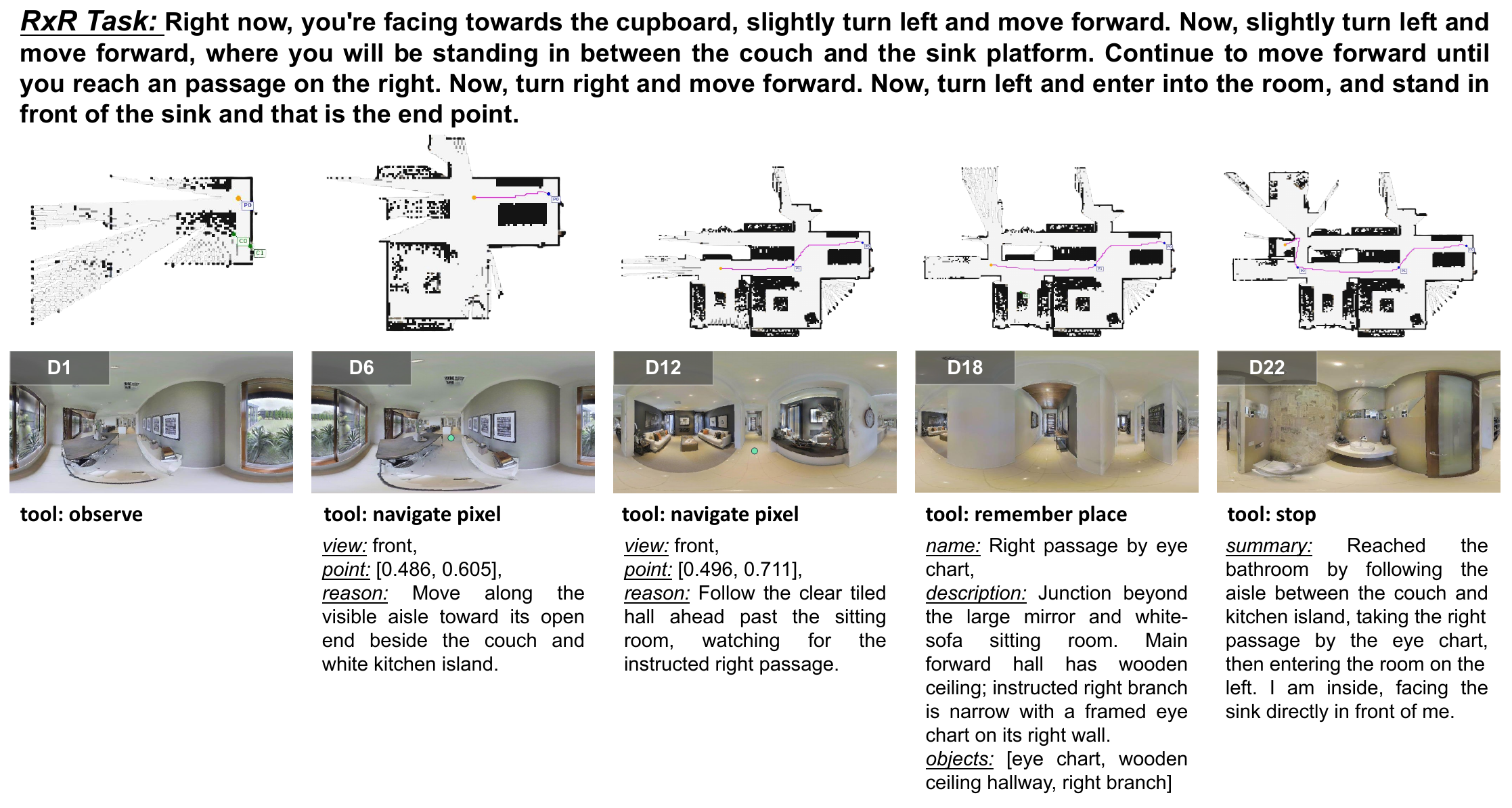}
\caption{\textbf{Pixel navigation and landmark recording on RxR-CE.}
The agent follows the passage between a couch and kitchen island, identifies
the instructed right branch, and enters the room containing the sink.
Pixel selections express local floor destinations, while a
\tool{remember\_place} request records the eye-chart junction and its
distinguishing geometry. The final observation shows the sink directly
ahead before the task is submitted. Orange dots on the maps indicate the agent's position after executing the
corresponding decision. $P_i$ denotes a saved place in persistent memory. In the panoramas, green markers indicate the selected waypoint or pixel. $D_i$ denotes the $i$-th tool-level decision.}
\label{fig:qual-rxr}
\end{figure}
\FloatBarrier

\figref{fig:qual-rxr} shows a detailed instruction being resolved through
successive local observations. The couch and white kitchen island establish
the first aisle; later views reveal the hallway and the required right
passage. At the
junction, it records ``Right passage by eye chart,'' retaining the framed
eye chart, narrow branch, and wooden-ceiling hallway as spatial cues.
Further requests select the right passage and then the bathroom entrance in
the left view. A local turn finally aligns the observation with the sink.
This example illustrates landmark recording during execution; it does not
require revisiting a saved place within the same task.

\clearpage
\subsection{HM3D v2: From a Visible Plant to an Approach Target}
\label{app:qual-hm3d}

\begin{figure}[!htbp]
\centering
\includegraphics[width=\linewidth]{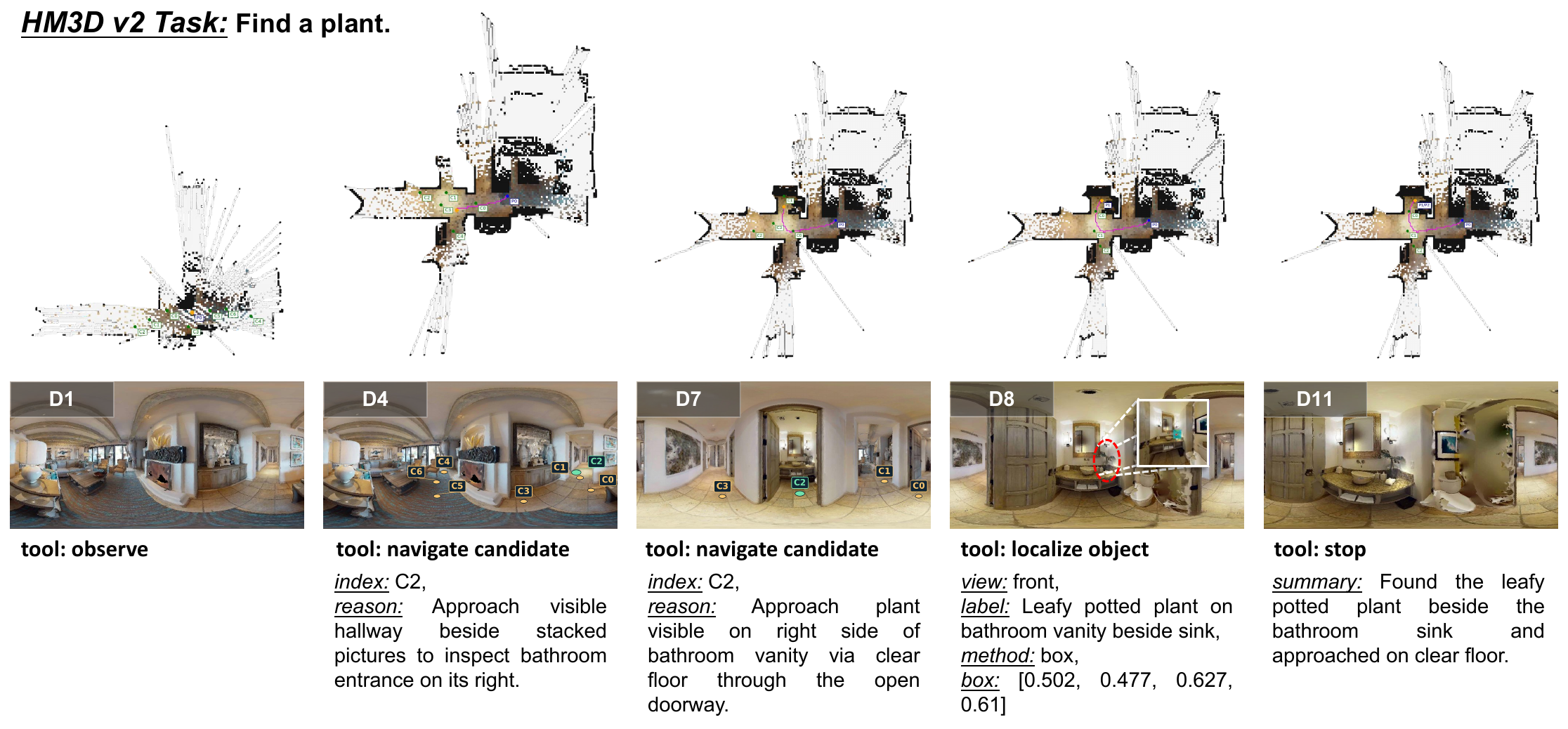}
\caption{\textbf{Object grounding on HM3D v2.}
Starting from a living area, the agent approaches the hallway and bathroom
entrance through visible free floor. Once the potted plant beside the vanity
is visible, \tool{localize\_object} selects its image region with a bounding
box. The subsequent approach ends beside the bathroom sink, where the agent
submits an explicit stopping decision. Orange dots on the maps indicate the agent's position after executing the
corresponding decision. $P_i$ denotes a saved place in persistent memory. In the panoramas, $C_i$ labels candidate waypoint $i$,
and green markers indicate the selected waypoint or pixel. $D_i$ denotes the $i$-th tool-level decision.}
\label{fig:qual-hm3d}
\end{figure}
\FloatBarrier

In \figref{fig:qual-hm3d}, a category-level goal becomes a specific visual
target as exploration reveals the bathroom. The first candidate request
uses the stacked pictures and open hallway as local references for reaching
a more informative viewpoint. From there, the agent identifies the leafy
plant on the right side of the vanity and chooses another candidate through
the open doorway. The bounding box in the subsequent
\tool{localize\_object} request specifies which part of the front view should
support RGB-D grounding. The recorded sequence then uses the resulting
estimate for object approach. The final panorama places the plant beside
the sink at close range. The example connects room-scale exploration,
image-based target selection, and geometric approach, with the agent
retaining responsibility for recognizing the plant and deciding when to
stop.

\clearpage
\subsection{HM3D-OVON: Turning a Landmark Description into Movement}
\label{app:qual-ovon}

\begin{figure}[!htbp]
\centering
\includegraphics[width=\linewidth]{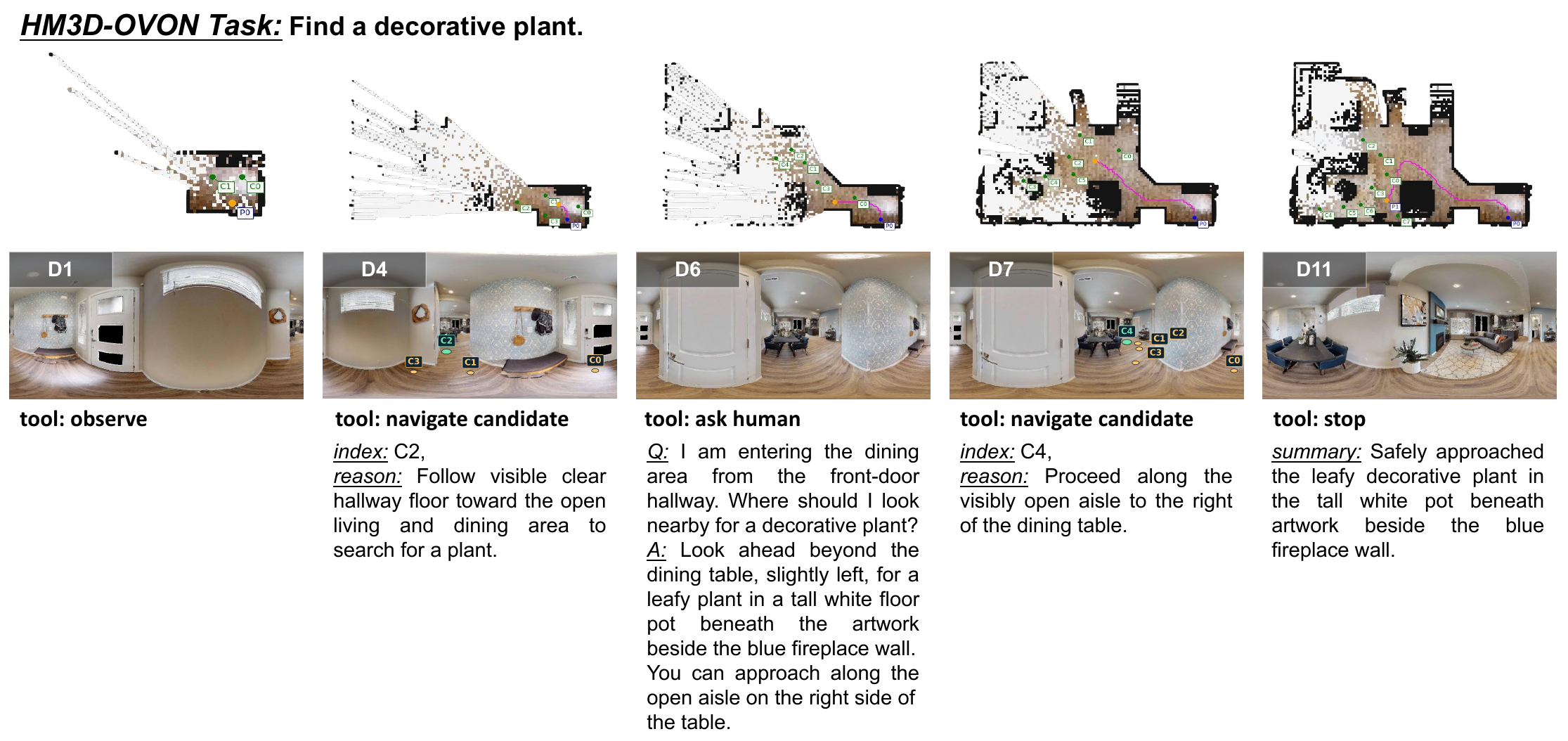}
\caption{\textbf{Active guidance during open-vocabulary object navigation.}
After entering the dining area, the agent asks where to look for a
decorative plant. The guide identifies a tall white pot beneath artwork
beside the blue fireplace wall and describes an aisle around the dining
table. A subsequent candidate request follows that aisle, and the final
view shows the agent beside the indicated plant. Orange dots on the maps indicate the agent's position after executing the
corresponding decision. $P_i$ denotes a saved place in persistent memory. In the panoramas, $C_i$ labels candidate waypoint $i$,
and green markers indicate the selected waypoint or pixel. $D_i$ denotes the $i$-th tool-level decision.}
\label{fig:qual-ovon}
\end{figure}
\FloatBarrier

\figref{fig:qual-ovon} shows a question changing the information available
for the next movement decision. The initial hallway observations provide
an accessible route toward the living and dining area but do not yet
establish a specific plant destination. The agent asks after reaching the
room entrance. The answer combines appearance cues, including the white
pot and artwork, with a route cue along the right side of the table.
The next candidate request follows this open aisle, and later observations
expose the plant beside the blue fireplace wall. In the complete record,
the agent also localizes the visible plant and executes an object-approach
request before stopping. The interaction thus supplies a semantic and
spatial reference that can be checked against new views and acted on through
the same motion interfaces used elsewhere in the framework.

\clearpage
\subsection{VL-LN: Identifying the Intended Table}
\label{app:qual-vlln}

\begin{figure}[!htbp]
\centering
\includegraphics[width=\linewidth]{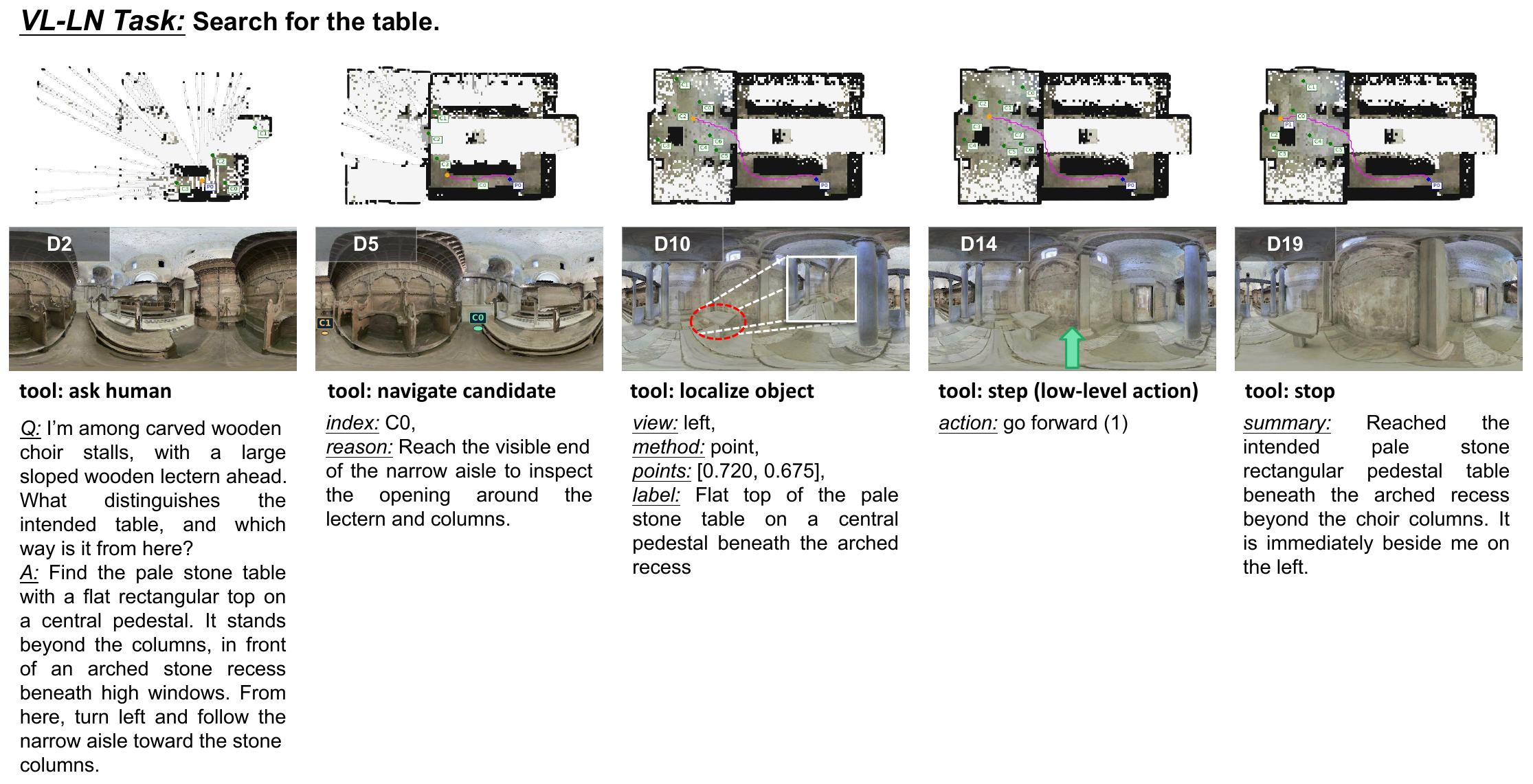}
\caption{\textbf{Instance disambiguation on VL-LN.}
The category-level request to find a table is clarified through a description
of a pale stone rectangular top on a central pedestal beneath an arched
recess. The agent follows the narrow aisle, selects a point on the observed
table for localization, and uses local movement to adjust its position.
Its final summary refers to the same structural and spatial attributes. Orange dots on the maps indicate the agent's position after executing the
corresponding decision. $P_i$ denotes a saved place in persistent memory. In the panoramas, $C_i$ labels candidate waypoint $i$,
and green markers indicate the selected waypoint or pixel. $D_i$ denotes the $i$-th tool-level decision.}
\label{fig:qual-vlln}
\end{figure}
\FloatBarrier

The initial view in \figref{fig:qual-vlln} contains carved wooden choir
stalls, a sloped lectern, and multiple stone columns. The short task text
does not specify which visible structure is the intended table, so the
agent asks for distinguishing properties and a direction from its current
viewpoint. The guide's answer supplies a material, shape, support structure,
and architectural context. A candidate request then advances along the
aisle to inspect the opening around the lectern and columns. After the pale
stone table appears, a point-based \tool{localize\_object} request selects its
flat top in the left view. The remaining recorded movements use candidate
and discrete requests to approach and adjust beside it. This sequence
separates target identification, geometric localization, and final
positioning while preserving a consistent instance description across
these decisions.

\clearpage
\subsection{CoIN: Correcting an Ambiguous Bedroom Choice}
\label{app:qual-coin}

\begin{figure}[!htbp]
\centering
\includegraphics[width=\linewidth]{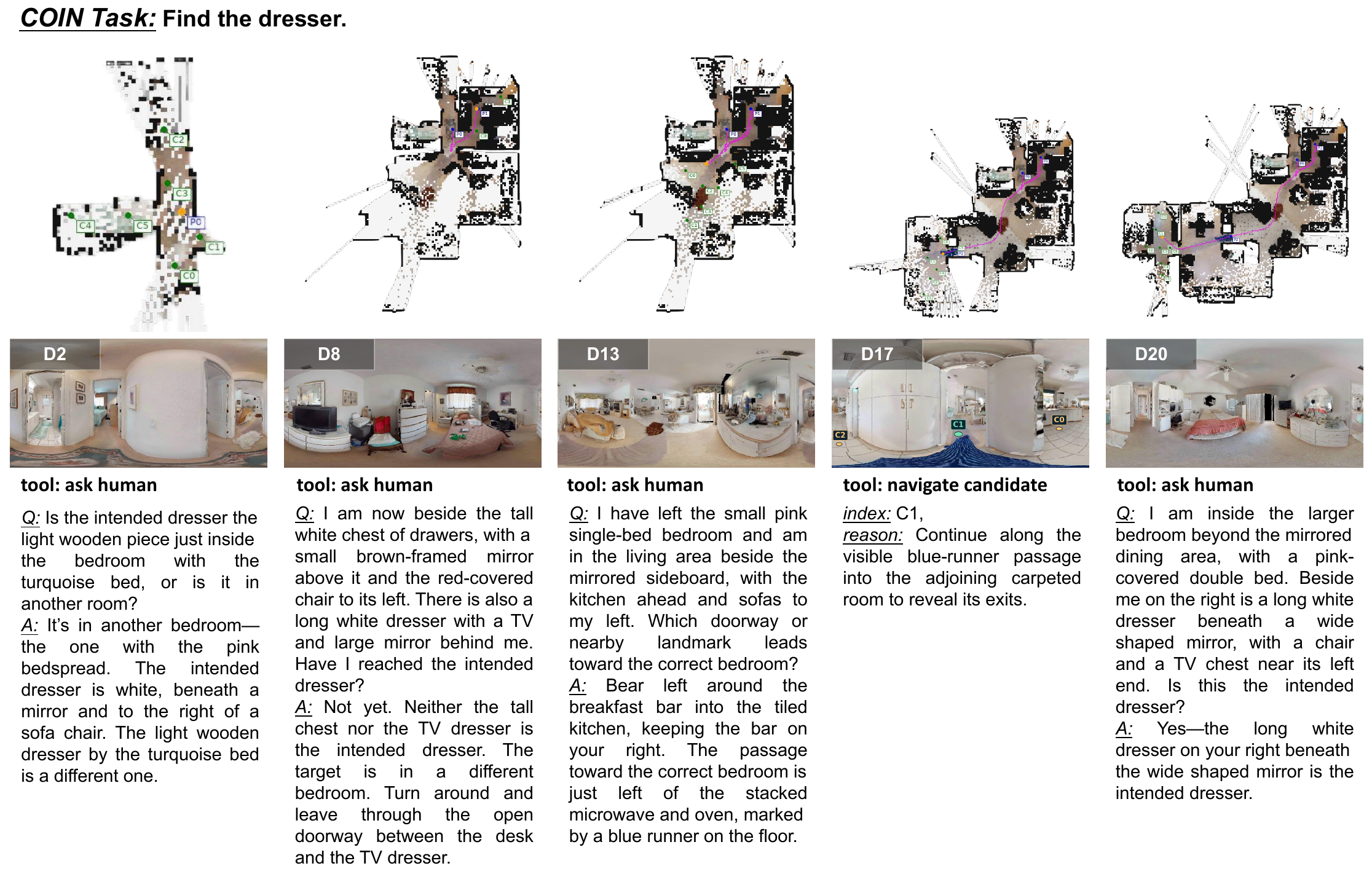}
\caption{\textbf{Iterative clarification on CoIN.}
The agent first distinguishes the requested dresser from one beside a
turquoise bed. A later question reveals that the furniture in a small
pink-bed bedroom is also incorrect. Further guidance identifies a
blue-runner passage through the kitchen. The final question verifies the
long white dresser beneath a wide mirror in the larger bedroom. Orange dots on the maps indicate the agent's position after executing the
corresponding decision. $P_i$ denotes a saved place in persistent memory. In the panoramas, $C_i$ labels candidate waypoint $i$,
and green markers indicate the selected waypoint or pixel. $D_i$ denotes the $i$-th tool-level decision.}
\label{fig:qual-coin}
\end{figure}
\FloatBarrier

\figref{fig:qual-coin} illustrates how an initial interaction may be insufficient to resolve target ambiguity, requiring further clarification through subsequent questions. The guide initially describes the target as a white dresser beneath a mirror in a bedroom with a pink bedspread. However, these attributes are insufficient to distinguish between the relevant bedrooms.
The agent then reaches a small bedroom with a pink bed and describes the nearby chest, mirror, chair, and TV dresser in a follow-up question. The guide explicitly rules out these candidates, prompting the agent to leave the room.
The agent subsequently asks for additional route guidance. The guide identifies a passage through the kitchen near the stacked appliances and a blue runner. Following this passage, the agent eventually confirms through interaction that the intended target is the long white dresser beneath a wide mirror in the larger bedroom.
This example demonstrates how active interaction and persistent memory jointly support the correction of plausible but incorrect spatial interpretations. Through successive questions, the agent progressively eliminates incorrect targets and retains the resulting corrections in memory to inform subsequent navigation decisions.

\clearpage
\subsection{LHPR-VLN: Reusing a Landmark across Ordered Subtasks}
\label{app:qual-lhpr}

\begin{figure}[!htbp]
\centering
\includegraphics[width=\linewidth]{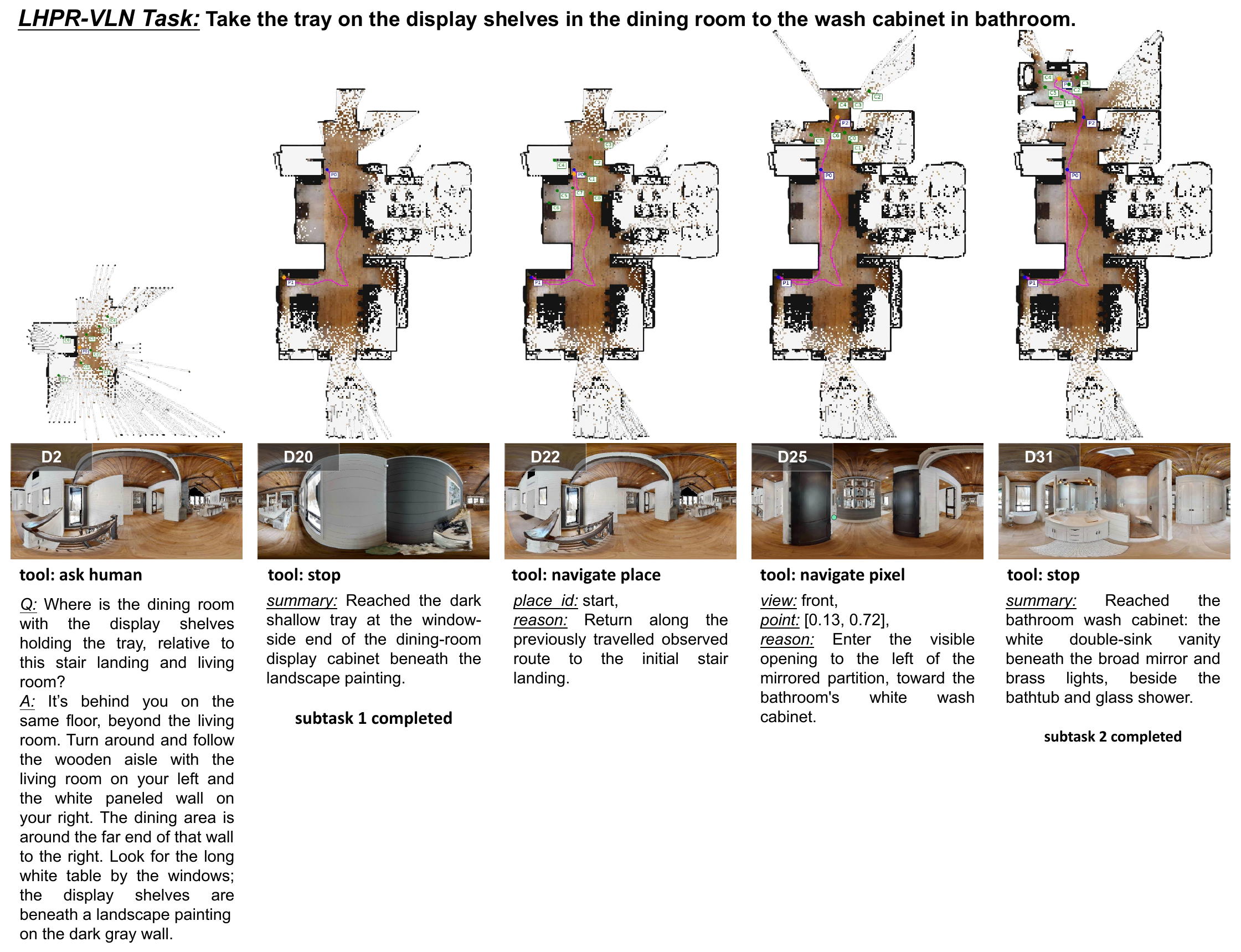}
\caption{\textbf{Continuous two-stage navigation on LHPR-VLN.}
The agent first reaches the dining-room tray and submits that stage with
\tool{stop}. It then calls \tool{navigate\_place} with the saved
\texttt{start} location to return toward the initial stair landing.
A pixel request selects the bathroom opening, and the recorded execution
continues with candidate movements before the final stop at the wash
cabinet. Both destinations are visited in one continuous sequence. Orange dots on the maps indicate the agent's position after executing the
corresponding decision. $P_i$ denotes a saved place in persistent memory. In the panoramas, green markers indicate the selected pixel. $D_i$ denotes the $i$-th tool-level decision.}
\label{fig:qual-lhpr}
\end{figure}
\FloatBarrier

\figref{fig:qual-lhpr} illustrates two navigation stages that share the observed map and place records.
In the first stage, the agent actively asks for guidance and relates the current stair-landing view to the dining-room destination using spatial cues such as the wooden aisle, white paneled wall, long table, and landscape painting. After reaching the tray, the agent records its endpoint and submits the first navigation stage.
In the second stage, new guidance indicates that the bathroom lies beyond the initial stair landing, prompting the agent to explicitly return to the previously saved \texttt{start} location. In this way, the agent reuses familiar spatial landmarks and historical memory to guide navigation toward a new destination.
Upon approaching the bathroom, the agent successfully enters the room through candidate navigation requests. The final view shows the target white vanity beneath the mirror.
The two stages are submitted through separate stopping decisions and completed in the specified order within a single continuous navigation sequence.

\clearpage
\subsection{GOAT-Bench: Maintaining Experience across Goal Modalities}
\label{app:qual-goat}

\begin{figure}[!htbp]
\centering
\includegraphics[width=\linewidth]{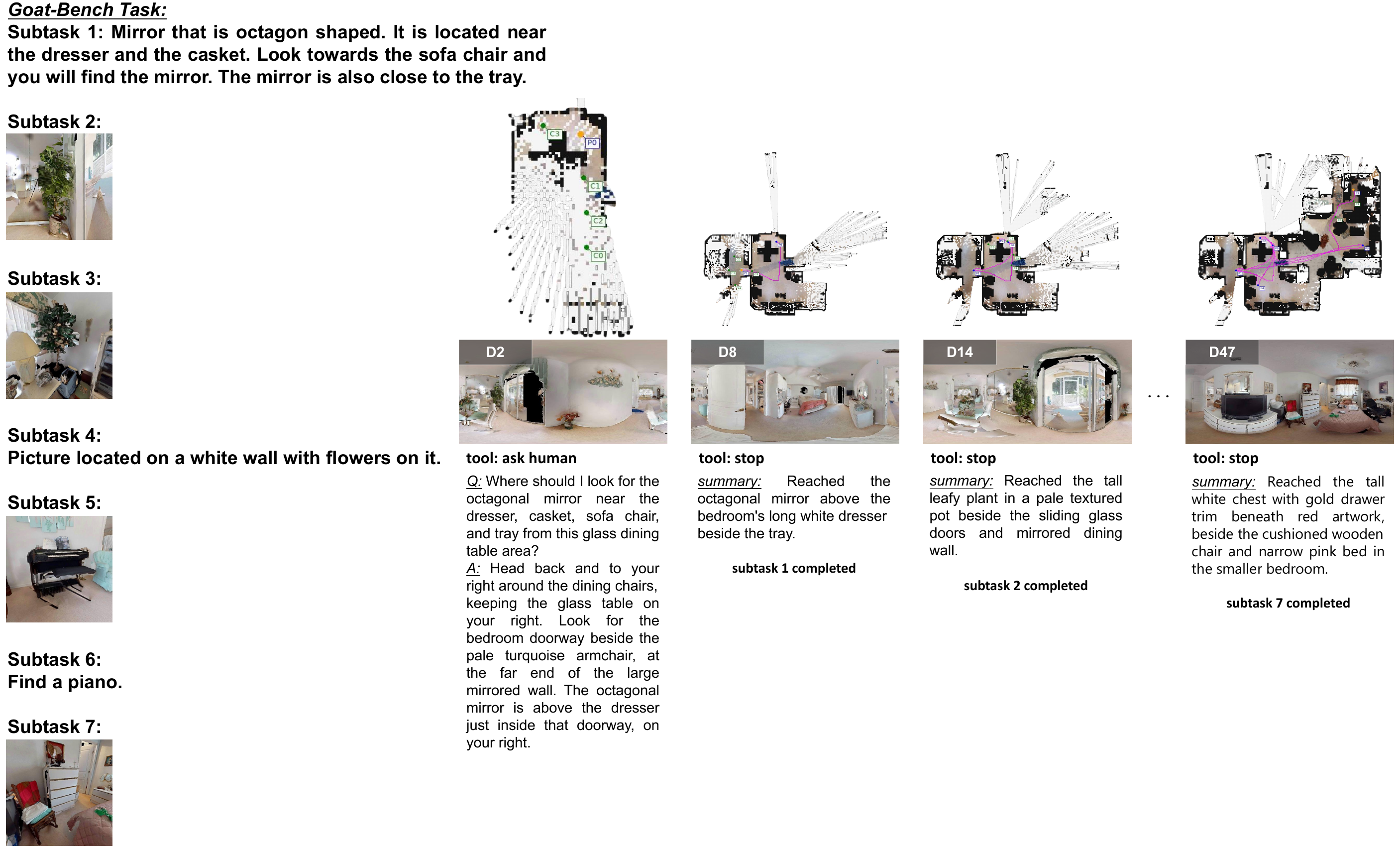}
\caption{\textbf{Selected stages of a seven-subtask GOAT-Bench sequence.}
The goals combine language descriptions, reference images, and an object
category. The execution panels show the initial mirror search, completion
of the image-specified plant subtask, and completion of the final chest
subtask. The observed map retains exploration across these stages.
Only subtasks 1, 2, and 7 are shown in the execution panels. Orange dots on the maps indicate the agent's position after executing the
corresponding decision. $P_i$ denotes a saved place in persistent memory. $D_i$ denotes the $i$-th tool-level decision.}
\label{fig:qual-goat}
\end{figure}
\FloatBarrier

\figref{fig:qual-goat} separates changes in the goal representation from
continuity of the environment. The first target is an octagonal mirror
specified through relations to nearby furniture; the next is a plant
specified by an image. The agent asks for the mirror's location, reaches
the indicated bedroom, and records the endpoint. During the plant subtask,
the recorded execution revisits \texttt{start} to return to the dining area
where that object had been visible. Later goals extend the same map and
place records. Reuse still requires checking current evidence: during the
final chest search, the agent initially returns to the remembered mirror
bedroom, asks whether it is the correct room, and receives a correction.
It then navigates through another remembered location toward the smaller
bedroom. The final frame shows the matching chest. Persistent memory thus
provides reusable spatial references while interaction can correct their
relevance to the current goal.

\FloatBarrier

\endgroup

\end{document}